\pdfoutput=1

\documentclass[11pt]{article}

\usepackage{acl} 
\usepackage{natbib}

\usepackage{times}
\usepackage{latexsym}
\usepackage{mdframed}
\usepackage{adjustbox}
\usepackage{booktabs}
\usepackage{multicol}
\usepackage{verbatim}
\usepackage{caption}
\usepackage{booktabs}
\usepackage{multirow}
\usepackage{colortbl}
\usepackage{xcolor}
\usepackage[inline]{enumitem}
\usepackage{array}
\usepackage{listings}
\usepackage{siunitx}
\usepackage{tikz}
\usepackage{amsmath}
\usepackage{tcolorbox}
\usepackage{graphicx}
\usepackage[utf8]{inputenc}
\usepackage{csquotes}
\usepackage[T1]{fontenc}
\usepackage{tcolorbox}
\usepackage{epigraph}
\definecolor{NavyBlue}{HTML}{001f3f}
\definecolor{SkyBlue}{HTML}{87CEEB}
\definecolor{RoyalBlue}{HTML}{4169E1}
\definecolor{LightSteelBlue}{HTML}{B0C4DE}
\definecolor{CobaltBlue}{HTML}{0047AB}
\definecolor{PowderBlue}{HTML}{B0E0E6}

\usepackage{microtype}

\usepackage{inconsolata}

\usepackage{graphicx}

\title{A Multi-Agent Framework for Mitigating Dialect Biases in Privacy Policy Question-Answering Systems}

\author{\DJ or\dj e Klisura$^1$, Astrid R Bernaga Torres$^2$, Anna Karen Gárate-Escamilla$^2$,\\\textbf{Rajesh Roshan Biswal$^2$, Ke Yang$^1$, Hilal Pataci$^1$, Anthony Rios$^1$} \\
  $^1$University of Texas at San Antonio \\
  $^2$Tecnológico de Monterrey \\
  \texttt{\{Dorde.Klisura, Anthony.Rios\}@utsa.edu} \\}

\begin{document}
\maketitle
\begin{abstract}
Privacy policies inform users about data collection and usage, yet their complexity limits accessibility for diverse populations. Existing Privacy Policy Question Answering (QA) systems exhibit performance disparities across English dialects, disadvantaging speakers of non-standard varieties. We propose a novel multi-agent framework inspired by human-centered design principles to mitigate dialectal biases. Our approach integrates a Dialect Agent, which translates queries into Standard American English (SAE) while preserving dialectal intent, and a Privacy Policy Agent, which refines predictions using domain expertise. Unlike prior approaches, our method does not require retraining or dialect-specific fine-tuning, making it broadly applicable across models and domains. Evaluated on PrivacyQA and PolicyQA, our framework improves GPT-4o-mini's zero-shot accuracy from 0.394 to 0.601 on PrivacyQA and from 0.352 to 0.464 on PolicyQA, surpassing or matching few-shot baselines without additional training data. These results highlight the effectiveness of structured agent collaboration in mitigating dialect biases and underscore the importance of designing NLP systems that account for linguistic diversity to ensure equitable access to privacy information.

\end{abstract}

\section{Introduction}

Privacy policies are essential documents that outline how organizations collect, use, and share personal data. Yet, their effectiveness is undermined by excessive length, legal complexity, and inaccessible language, making it difficult for users to understand their rights and risks \cite{ravichander2019question, ahmad2020policyqa}. Privacy Policy Question Answering (QA) systems aim to bridge this gap by providing users with concise, query-driven insights. However, existing systems remain largely indifferent to linguistic diversity, particularly the nuanced variations in English dialects, thereby constraining equitable access to privacy information. This oversight is especially consequential in real-world deployments, where dialectal differences fundamentally shape how users parse and interpret complex legal and technical content. Specifically, the Electronic Privacy Information Center (EPIC) states the following on their website~\cite{EPIC_Privacy_Racial_Justice}:
\begin{figure}
    \centering
    \includegraphics[width=0.8\linewidth]{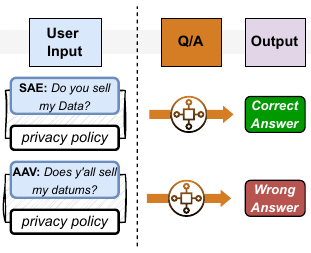}
    \caption{ Illustration of dialect-based disparities in Privacy Question Answering (QA). The QA model correctly answers a query phrased in Standard American English (SAE) but produces an incorrect response when the same query is asked in African American Vernacular English (AAVE).} \vspace{-0em}
    \label{fig:overview}
\end{figure}
\begin{quote}
    \textit{Marginalized communities are disproportionately harmed by data collection practices and privacy abuses from the both the government and private sector. Communities of color are especially targeted, discriminated against, and exploited through surveillance, policing, and algorithmic bias.}\\
    - EPIC
\end{quote}
From a privacy QA perspective, if all groups cannot ask questions to help protect their information effectively, those groups are at risk. We illustrate this issue in Figure~\ref{fig:overview}.

The challenge of dialectal bias in NLP has been extensively documented, with non-standard dialects such as African American Vernacular English (AAVE), Chicano English, and Aboriginal English often receiving subpar performance compared to Standard American English (SAE)~\cite{ziems2023multivalue, blodgett2018aaedisparities}. This disparity disproportionately affects marginalized communities, amplifying existing inequities and limiting access to language technologies for non-dominant speakers~\cite{sap2019socialbias, davidson2019hatebias}. While frameworks like Multi-VALUE have been developed to evaluate and mitigate dialect biases in general NLP tasks \cite{ziems2023multivalue}, no work has explored how such biases manifest in domain-specific applications like privacy policy QA.

Furthermore, much of the recent work on question-answering has focused on large language models (LLMs) and, in particular, prompting-based methods~\cite{lee2022type,yu2023exploring}. These systems are developed to work well generally for a wide audience. However, they struggle with geographical/cultural~\cite{lwowski2022measuring,liu2024culturally,naous-etal-2024-beer} and dialectal biases~\cite{lwowski2021risk,faisal-etal-2024-dialectbench} when used by specific communities. Hence, a fundamental question is, ``How can we tune the prompting procedures of LLMs to perform well for minority communities/dialects without collecting large amounts of training data from these communities to fine-tune models, which may be difficult, particularly in sensitive application domains?''


To address these limitations, we introduce a novel multi-agent\footnote{\textcolor{black}{We use the term \textit{multi-agent} to describe structured prompt-based collaboration between distinct roles invoked via large language models, rather than autonomous agents in classical multi-agent systems.}} collaboration framework for dialect-sensitive privacy policy QA. Our method integrates two specialized agents: a Dialect Agent and a Privacy Policy Agent. The Dialect Agent processes user queries in diverse dialects by translating them into SAE, providing relevant judgments, and explaining their reasoning. The Privacy Policy Agent further refines these outputs by leveraging domain-specific expertise to validate and improve predictions. This collaborative design allows us to mitigate dialectal biases without requiring task-specific retraining or extensive dialectal datasets, addressing the scalability challenges of previous approaches.

We evaluate our framework on the PrivacyQA and PolicyQA datasets, which include queries across a wide range of dialects generated using the Multi-VALUE framework. Our method significantly improves fairness and accuracy, reducing performance disparities across dialects by up to 82\% as measured by the maximum difference in F1 scores between dialects.  Furthermore, our approach achieves state-of-the-art performance in privacy policy QA, highlighting its robustness, scalability, and real-world applicability in mitigating dialectal biases while enhancing accessibility to critical privacy information.
Overall, we make the following contributions in this paper:
\begin{itemize}
    \item We perform an exhaustive benchmark of dialect biases for state-of-the-art LLMs applied to privacy question-answering datasets.
    \item We introduce a novel multi-agent framework that introduces direct knowledge about the dialect and/or minority group to mitigate biases and improve overall performance.
    \item We perform a comprehensive ablation and error analysis. Moreover, we provide implications for deploying this approach in practice.
\end{itemize}

\section{Related Work}

\paragraph{NLP and Privacy.} 
NLP research in privacy policy extends beyond QA, tackling the structural and interpretive challenges of privacy policies.  To address this, various datasets have been developed to facilitate privacy policy research \cite{wilson2016creation,ramanath2014unsupervised, srinath2021privacy, amos2021privacy, manandhar2022smart}. Notable efforts include OPP-115, which focuses on classifying privacy practices within policies~\cite{chi2023plue}. Similarly, PolicyIE enables semantic parsing by identifying intents and filling slots related to privacy practices~\cite{ahmad2021policyie}. Named Entity Recognition (NER) tasks, such as PI-Extract, identify specific data types mentioned in privacy policies, supporting better automatic understanding~\cite{bui2021piextract}. The PLUE benchmark consolidates these tasks, providing a comprehensive evaluation framework for privacy policy language understanding~\cite{chi2023plue}. These initiatives have broadened the scope of privacy policy NLP by addressing tasks like classification, semantic parsing, and NER, creating a foundation for advanced applications in this domain.

Privacy policy QA has emerged as a critical area of study, aiming to streamline user interactions with these documents by retrieving concise and relevant answers to user queries. PrivacyQA introduced a sentence-level evidence retrieval framework, highlighting the inherent challenges of answerability and relevance~\cite{ravichander2019question}. PolicyQA advanced this approach by framing the task as span extraction, emphasizing the need for short and precise answers to improve accessibility~\cite{ahmad2020policyqa}. PLUE expanded the evaluation framework to include QA as one of its core tasks, demonstrating the value of domain-specific pre-training in improving QA accuracy~\cite{chi2023plue}. Despite significant progress, open challenges persist, particularly in addressing ambiguities, improving robustness to linguistic diversity, and ensuring fairness across user demographics, \textcolor{black}{as well as mitigating emerging security concerns in deploying large language models}~\cite{klisura}.

\paragraph{Dialectal NLP.} Dialect NLP research highlights significant performance disparities between dominant dialects, such as standard American English (SAE), and lower-resource dialects such as African American Vernacular English (AAVE), Chicano English and Indian English, raising concerns about fairness and equity in language technology~\cite{ziems2023multivalue, blodgett2018aaedisparities, jurgens2017languageid}. These disparities, evident in tasks such as dependency analysis, sentiment analysis, and hate speech detection, disproportionately affect marginalized communities~\cite{sap2019socialbias, davidson2019hatebias, jorgensen2016learning}. The lack of robust dialectal evaluation frameworks exacerbates these issues, reinforcing existing power imbalances in NLP systems~\cite{bender2021allocationalharms, hovy2016nlpequity}. Existing work, such as Multi-VALUE, addresses these gaps by creating rule-based perturbations and stress tests to evaluate model robustness across 50 English dialects~\cite{ziems2023multivalue, kortmann2020ewave}. Frameworks like DADA and TADA employ modular and task-agnostic approaches, enabling fine-grained adaptation and cross-dialectal robustness without requiring extensive task-specific data~\cite{liu2023dada, held2023tada}. These advancements are complemented by efforts to incorporate sociolinguistic insights into model development, addressing morphosyntactic variations and promoting scalable, equitable solutions for dialectal NLP~\cite{sun2023chinesedialects, demszky2019analyzing}. Together, these approaches underscore the critical need for inclusive NLP systems that mitigate dialectal biases and ensure equitable access to language technologies~\cite{blodgett2018aaedisparities, sap2019socialbias, davidson2019hatebias}. This paper uses the Multi-Value dialectal testing framework to evaluate biases in privacy QA tasks. Moreover, we overcome some of the limitations of prior dialectal technologies that require dialect-aware training frameworks~\cite{liu2023dada, held2023tada}. Instead, our framework only requires some initial (minimal) dialect information supplied as a prompt, minimizing some of the complexities in implementing prior work.

\paragraph{Multi-agent Modeling.} Multi-agent systems (MAS) have become increasingly prominent in NLP for coordinating specialized agents to handle complex and large-scale tasks. LongAgent~\cite{zhao2024longagent} addresses long-document QA by distributing text across agents and using iterative communication to reduce hallucinations and ensure consistent answers. Recent MAS work has also emphasized collective decision-making (CDM), with systems like GEDI~\cite{zhao2024gedi} applying voting methods such as ranked pairs and plurality to improve fairness and robustness. Beyond QA, MAS have proven effective in multi-turn reasoning~\cite{chen2023multiagentreasoning}, knowledge retrieval~\cite{liu2023knowledgeagents}, and structured prediction~\cite{xu2023structureddecision}, showcasing their versatility. These frameworks highlight how inter-agent collaboration and feedback loops can enhance performance, reliability, and inclusivity in a range of NLP applications.

\paragraph{LLM-based multi-agent systems.} \textcolor{black}{Recent work has explored LLM-based multi-agent systems that differ from classical approaches by coordinating agents through natural language rather than fixed protocols~\cite{li2024survey}. These systems assign roles like planner, critic, or explainer to individual models and enable them to collaborate via structured, prompt-based dialogue. Frameworks like CAMEL~\cite{li2023camel}, AutoAgents~\cite{chen2024autoagentsframeworkautomaticagent}, and ChatDev~\cite{qian2023chatdev} show how role-based agents can dynamically negotiate, critique, and refine their outputs to complete complex tasks like software development, multi-hop reasoning, or policy interpretation. While classical MAS emphasized distributed algorithms and communication protocols, LLM-based systems focus on emergent cooperation through language, enabling more flexible task decomposition and iterative problem-solving. Our work builds on this paradigm by prompting specialized agents (the Dialect and Privacy Policy agents) to engage in structured collaboration through role-specific prompting and iterative refinement.} 

\begin{figure*}[t]
    \centering
    \includegraphics[width=0.9\linewidth]{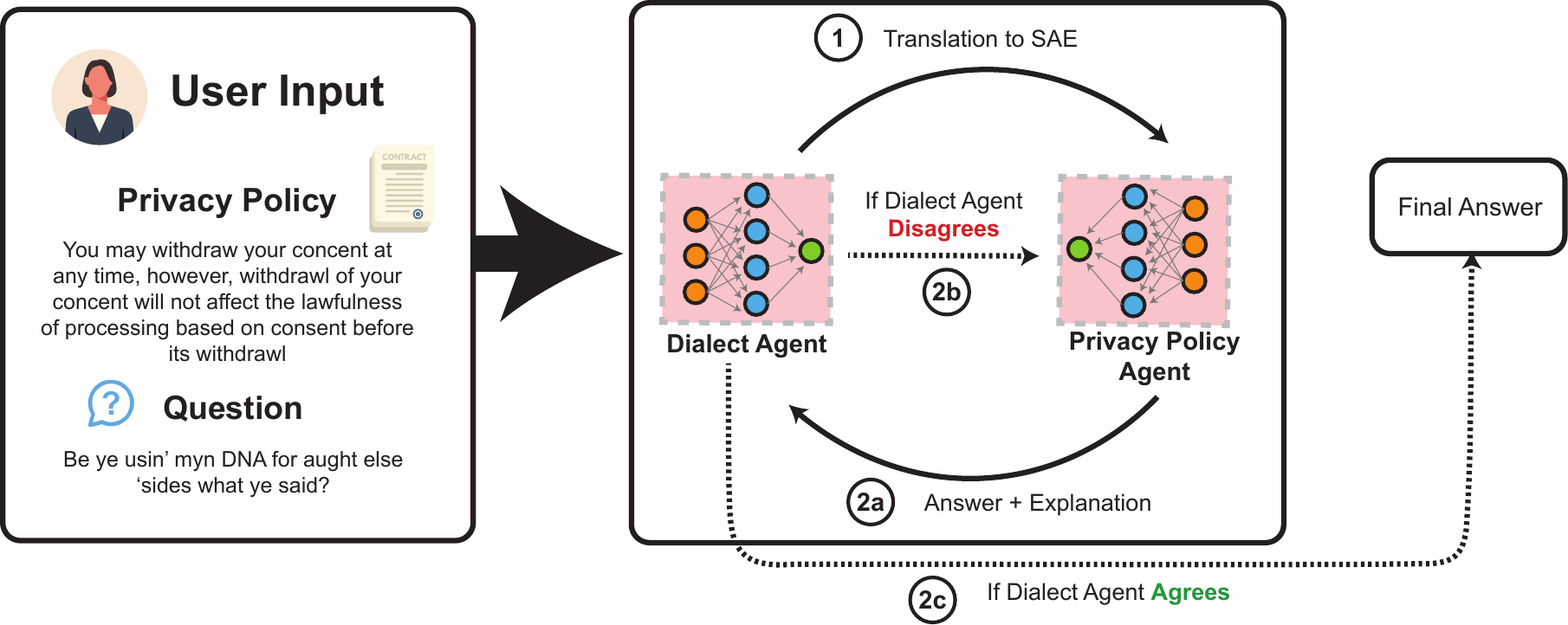}
    \caption{Our multi-agent framework for mitigating dialect biases in privacy QA. The Dialect Agent translates queries into Standard American English (SAE) and validates responses. The Privacy Policy Agent generates answers based on policy text. Disagreements trigger refinement, ensuring accurate and inclusive responses across dialects.}
    \label{fig:enter-label}
\end{figure*}

\section{Methodology}
Our primary objective is to reduce performance disparities in privacy policy QA across multiple large language models when queries are posed in diverse English dialects. To formally define the task, let \(q_d\) be a question in dialect \(d \in \mathcal{D}\) and let \(p\) be a corresponding privacy policy snippet. A QA model \(f\) produces an answer \(A = f(p, q_d)\), which is compared to a ground-truth answer \(A^*\). We measure correctness using a metric \(\Phi\). For a given dialect \(d\), the average performance of \(f\) is denoted by \(\Phi_{d}(f)\). We define the overall performance disparity \(\Delta(f)\) as:
\[
\Delta(f) = \max_{d_i, d_j \in \mathcal{D}} \bigl|\Phi_{d_i}(f) - \Phi_{d_j}(f)\bigr|.
\]
The goal is to design a QA framework \(F\) that minimizes \(\Delta(f)\) while maintaining average accuracy on privacy policy questions.

To achieve this, we introduce a multi-agent collaboration framework. Figure~\ref{fig:enter-label} provides a high-level overview of our approach. The framework mirrors a human-centered design~\cite{cooley2000human} approach by prioritizing usability, fairness, and inclusivity in PrivacyQA systems. It leverages two specialized agents: a Dialect Agent and a Privacy Agent, designed to adapt to user needs and linguistic diversity. The Dialect Agent is an intermediary that translates ``non-standard'' dialect questions into SAE while preserving the user's query's original intent and cultural nuances. This, again, is based on human-centered design, where we try to add user information about the dialect they speak to the model to improve performance. This ensures that speakers of diverse dialects are not disadvantaged when interacting with privacy policy information because they are explicitly addressed in the model.

Meanwhile, the Privacy Agent interprets privacy policy segments\footnote{Privacy policies typically encompass ten major categories of data practices. These include First Party Collection (FP),
Third Party Sharing/Collection (TP), Data Retention (DR), and Data Security (DS), which
explain how and why first and third parties collect, process, store, share, and protect customer data. User rights are addressed through categories like User Choice/Control (UCC), User Access, Edit, Deletion (UAED), and Do Not Track (DNT)\cite{wilson2016creation}.} and generates accurate, policy-oriented answers that remain accessible and relevant across different linguistic backgrounds. By structuring the system as a collaborative process that integrates dialect-aware adaptation (from the Dialect Agent) and domain expertise (from the Privacy Agent), our approach embodies human-centered design principles---ensuring adequate performance on dialects beyond SAE.  We describe the agents below.

\vspace{2mm} \noindent \textbf{Step 1: Dialect Agent.} The Dialect Agent is prompted to act as an expert in diverse English dialects. Before processing any user query, it is given a concise yet detailed summary of a particular dialect’s key linguistic properties, including (very brief) phonetic, grammatical, lexical, and cultural aspects. Please see Appendix~\ref{sec:dialect} with examples. This setup enables the Dialect Agent to translate a user's dialectal question into SAE accurately and, subsequently, to validate whether the final answer aligns with the user’s original intent.

When a user provides a privacy policy segment and a question in a non-standard dialect, the question first goes to the Dialect Agent. Its task is to translate the query into clearly understandable SAE using its background knowledge about the dialect. Specifically, it is provided with the following prompt:\footnote{The prompts have been somewhat abbreviated for space considerations. See Appendix~\ref{appendix:agent-prompts} for full versions.} 
\begin{tcolorbox}[colback=gray!5!white, colframe=black, fontupper=\small, width=\linewidth, boxsep=0pt, title=Prompt] \emph{ ``You are an expert linguist specializing in the following dialect: $\{dialect\_info\}$.
Your task is to translate the following question from this dialect into clear, standard American English. Ensure that the translation is easily understandable to a general audience.
'' } \end{tcolorbox}
\noindent where $dialect\_info$ is the dialect information for that particular dialect. The output of this step is a standardized version of the user’s question, ready to be processed by the Privacy Agent.

\vspace{2mm} \noindent \textbf{Step 2a: Privacy Agent.} Once the dialectal query has been translated to SAE, it is handed over to the Privacy Agent along with the relevant segment of the privacy policy. The Privacy Agent is prompted as a domain expert, possessing comprehensive knowledge of typical privacy policy structures and terminologies.

The Privacy Agent uses the translated question and the given policy snippet to craft an initial response. The focus is on extracting accurate, succinct information from the policy segment that addresses the user’s query. The general prompt looks as follows:
\begin{tcolorbox}[colback=gray!5!white, colframe=black, fontupper=\small, width=\linewidth, boxsep=0pt, title=Prompt] \emph{ ``You are a privacy policy expert. Review the provided policy segment and answer the following question in a concise manner, ensuring factual accuracy. Base your response solely on the information in the policy segment.'' } \end{tcolorbox}
\noindent The Privacy Agent outputs both the initial answer and a brief rationale, indicating how the policy text justifies that answer.

\vspace{2mm} \noindent \textbf{Steps 2b and 2c: Evaluation by Dialect Agent}. Next, we provide the dialect agent with the original dialectal question, the policy segment, and the Privacy Agent’s proposed answer to the Dialect Agent. The Dialect Agent then evaluates whether the answer sufficiently captures the user’s intent and does not overlook subtle dialect-specific nuances. To do this, we provide the dialect agent the following prompt:
\begin{tcolorbox}[colback=gray!5!white, colframe=black, fontupper=\small, width=\linewidth, boxsep=0pt, title=Prompt] \emph{ ``Based on your understanding of the dialect’s linguistic and cultural nuances, determine whether the Privacy Agent’s answer fully addresses the user’s original question. Are there any discrepancies or misunderstandings that arise from the dialectal phrasing?'' }
\end{tcolorbox}
\noindent If the Dialect Agent confirms the answer is satisfactory, this output is accepted as final and step 2c is followed to return the final answer. If it flags potential inaccuracies or misunderstandings (for instance, the Privacy Agent missed the user’s intended meaning due to unique dialectal expressions), the process moves into a reconsideration stage (Step 2b) instead.

Upon receiving negative feedback from the Dialect Agent, the Privacy Agent revisits its initial answer. It is prompted to update or refine its response based on the Dialect Agent’s observations regarding the original question’s intent. The prompt is defined as follows:
\begin{tcolorbox}[colback=gray!5!white, colframe=black, fontupper=\small, width=\linewidth, boxsep=0pt, title=Prompt] \emph{ ``You received feedback indicating that certain elements of the user’s dialectal query were not fully addressed. Please revise your previous answer to incorporate the Dialect Agent’s insights and ensure the user’s intent is accurately captured.'' } \end{tcolorbox}
\noindent The Privacy Agent will then return another answer and rationale to the Dialect Agent. We will repeat this process until the agreement is met or a maximum number of iterations is met (we only loop a maximum of 2 times). This loop ensures that dialect nuances are not lost while improving the correctness of policy-based answers. {Note that in few-shot settings, we use a total of 8 examples per prompt for each agent. These examples reflect diverse dialects, question types, and policy scenarios, helping the agents generalize across linguistic and contextual variation.

\begin{table}[t]
\centering
\small
\resizebox{0.85\linewidth}{!}{%
\begin{tabular}{lrr}
\toprule
 & \textbf{PrivacyQA} & \textbf{PolicyQA} \\ 
& \textbf{Mobile Apps} & \textbf{Websites} \\ \midrule
\textbf{\# Policies} & 35 & 115 \\ 
\textbf{\# Questions} & 1,750 & 714 \\ 
\textbf{\# Annotations} & 3,500 & 25,017 \\ \bottomrule
\end{tabular}}
\caption{Statistics for Privacy Policy QA datasets.}
\label{tab:privacy_policy_datasets_2}
\end{table}

\begin{table*}[t]
\centering
\small
\resizebox{\textwidth}{!}{%
\begin{tabular}{lr|rrrrrrrr}
\toprule
\cmidrule(lr){2-8}
 \textbf{Model} & \textbf{SAE ($\uparrow$)} & \textbf{RAAVE ($\uparrow$)} & \textbf{Jamaican ($\uparrow$)} & \textbf{Aboriginal ($\uparrow$)} & \textbf{Welsh ($\uparrow$)} & \textbf{SWE ($\uparrow$)} & \textbf{AVG ($\uparrow$)} & \textbf{AVG Diff ($\downarrow$)} & \textbf{Max Diff ($\downarrow$)} \\
\midrule
GPT-4o-mini Zero               & .394 & .344 & .332 & .329 & .312 & .301 & .335 & .022 & .093 \\
GPT-4o-mini Few                & .605 & .573 & .562 & .555 & .547 & .547 & .565 & .016 & .058 \\
GPT-4o-mini Multi-agent-zero (ours)     & \textbf{.601} & \textbf{.588} & \textbf{.578} & \textbf{.587} & \textbf{.592} & \textbf{.576} & \textbf{.587} & \textbf{.007} & \textbf{.025} \\
GPT-4o-mini Multi-agent-few (ours)     & \textbf{.611} & \textbf{.595} & \textbf{.596} & \textbf{.602} & \textbf{.592} & \textbf{.594} & \textbf{.598} & \textbf{.005} & \textbf{.019} \\ \midrule
Llama 3.1 Zero           & .469 & .349 & .370 & .325 & .356 & .336 & .368 & .035 & .144 \\
Llama 3.1 Few            & .546 & .463 & .469 & .448 & .485 & .446 & .476 & .026 & .100 \\
Llama 3.1 Multi-agent-zero (ours)  & \textbf{.549} & \textbf{.527} & \textbf{.520 }& \textbf{.524} & \textbf{.523 }& \textbf{.526} & \textbf{.528 }& \textbf{.007} & \textbf{.029} \\
Llama 3.1 Multi-agent-few (ours)   & \textbf{.555} & \textbf{.525} & \textbf{.523} & \textbf{.529} & \textbf{.522 }& \textbf{.528} & \textbf{.530} & \textbf{.008} & \textbf{.033 }\\ \midrule 
DeepSeek-R1 Zero            & .532 & .510 & .547 & .529 & .532 & .512 & .527 & .011 & .037 \\
DeepSeek-R1 Few             & .581 & .549 & .547 & .517 & .556 & .541 & .549 & .014 & .064 \\
DeepSeek-R1 Multi-agent-zero (ours)  & \textbf{.582 }& \textbf{.579} & \textbf{.583} & \textbf{.579 }& .\textbf{566} & \textbf{.573 }& \textbf{.577} & \textbf{.005 }& \textbf{.017} \\
DeepSeek-R1 Multi-agent-few (ours)   & .533 & \textbf{.606} & \textbf{.585 }& \textbf{.581} & \textbf{.557 }& \textbf{.569} & \textbf{.572 }& .019 & .073 \\
\bottomrule
\end{tabular}}
\caption{Performance comparison on PrivacyQA across dialects. Our multi-agent framework (bold) improves accuracy and reduces disparities (AVG Diff and Max Diff) compared to baseline models (GPT-4o-mini, Llama 3.1, and DeepSeek-R1). Results are shown for Standard American English (SAE), Rural African American Vernacular English (RAAV), Jamaican English, Aboriginal English, Welsh English, and Southwest England Dialect (SWE).}
\label{tab:empty-table-1}
\end{table*}

\begin{table*}[t]
\centering
\small
\resizebox{\textwidth}{!}{%
\begin{tabular}{lr|rrrrrrrrr}
\toprule
\cmidrule(lr){2-8}
 \textbf{Model} & \textbf{SAE ($\uparrow$)} & \textbf{RAAVE ($\uparrow$)} & \textbf{Jamaican ($\uparrow$)} & \textbf{Aboriginal ($\uparrow$)} & \textbf{Welsh ($\uparrow$)} & \textbf{SWE ($\uparrow$)} & \textbf{AVG ($\uparrow$)} & \textbf{AVG Diff ($\downarrow$)} & \textbf{Max Diff ($\downarrow$)} \\ \midrule
GPT-4o-mini Zero               & .352 & .343 & .332 & .338 & .331 & .323 & .337 & .008 & .029 \\
GPT-4o-mini Few                & .478 & .423 & .458 & .452 & .444 & .438 & .449 & .014 & .055 \\
GPT-4o-mini Multi-agent-zero (ours)  & \textbf{.464 }&\textbf{ .444} & \textbf{.451} & \textbf{.458} & \textbf{.447} & \textbf{.445 }& \textbf{.452} & \textbf{.006 }& \textbf{.020 }\\
GPT-4o-mini Multi-agent-few (ours)   & \textbf{.484 }& \textbf{.460} & \textbf{.475} & \textbf{.473} & \textbf{.469} & \textbf{.467} & \textbf{.471} & \textbf{.006 }& \textbf{.024 }\\ \midrule
Llama 3.1 Zero           & .310 & .260 & .268 & .231 & .237 & .289 & .266 & .023 & .079 \\
Llama 3.1 Few            & .412 & .332 & .360 & .357 & .393 & .370 & .371 & .021 & .080 \\
Llama 3.1 Multi-agent-zero (ours)  &\textbf{ .381 }& \textbf{.374} &\textbf{ .368 }& \textbf{.358} & \textbf{.372} & \textbf{.368} & \textbf{.370 }& \textbf{.006 }& \textbf{.023 }\\
Llama 3.1 Multi-agent-few (ours)   & .400 & \textbf{.380 }& \textbf{.391} & \textbf{.385} & \textbf{.394 }& \textbf{.372 }& \textbf{.387 }& \textbf{.008 }& \textbf{.028} \\ \midrule
DeepSeek-R1 Zero            & .455 & .436 & .429 & .437 & .422 & .422 & .434 & .009 & .033 \\
DeepSeek-R1 Few             & .446 & .483 & .468 & .472 & .492 & .477 & .473 & .011 & .046 \\
DeepSeek-R1 Multi-agent-zero (ours)  & .451 & \textbf{.480 }& \textbf{.474 }& \textbf{.483} & \textbf{.463 }& \textbf{.481} & \textbf{.472 }& .010 & \textbf{.032} \\
DeepSeek-R1 Multi-agent-few (ours)   & \textbf{.474} & .476 & \textbf{.494 }& \textbf{.480} & .487 & \textbf{.480 }& \textbf{.482} & \textbf{.006} & \textbf{.020 }\\
\bottomrule
\end{tabular}}
\caption{Performance comparison on PolicyQA across dialects. Our multi-agent framework (bold) improves accuracy and reduces disparities (AVG Diff and Max Diff) compared to baseline models (GPT-4o-mini, Llama 3.1, and DeepSeek-R1). Results are shown for Standard American English (SAE), Rural African American Vernacular English (RAAV), Jamaican English, Aboriginal English, Welsh English, and Southwest England Dialect (SWE).}
\label{tab:empty-table-2}
\end{table*}

\section{Evaluation}

\vspace{2mm} \noindent \textbf{Data.} We use two privacy QA datasets: PrivacyQA and PolicyQA. We provide the dataset statistics in Table~\ref{tab:privacy_policy_datasets_2} for complete details. \textbf{PrivacyQA}~\cite{ravichander2019question} is a dataset designed for answer sentence selection on mobile app privacy policies. It contains 1,750 privacy-related questions with over 3,500 expert-annotated answers from 35 policies. Given a question and a set of possible answers (sentences from the policy), a model must determine which, if any, correctly answers the question. Specifically, each answer candidate is classified as ``correct'' or ``incorrect.'' The dataset includes answerable and unanswerable questions, reflecting real-world challenges in understanding privacy policies. For example, for the question ``Will my data be sold to advertisers?'', a model must determine if the sentence ``We do not sell your personal information.'' is a valid answer.

\textbf{PolicyQA}~\cite{ahmad2020policyqa} is a dataset for question answering (QA) on website privacy policies. It includes 25,017 question-answer pairs from 115 privacy policies, helping users find clear answers to privacy-related questions. Instead of returning long text passages, PolicyQA provides short, precise answers. For example, given the question ``Is my information shared with others?'', the dataset might provide the answer ``We do not give that business your name and address.'' This makes it easier for users to find the information they need quickly.

We use the Multi-VALUE~\cite{ziems2023multivalue} framework to translate both PrivacyQA and PolicyQA into the dialects it supports (e.g., African American Vernacular English). The Multi-VALUE framework is a rule-based translation system designed to enhance cross-dialectal NLP by systematically transforming SAE into synthetic forms of 50 different English dialects. It applies 189 linguistic perturbation rules informed by dialectology research to modify syntax and morphology while preserving semantics, enabling stress testing and data augmentation for NLP models. In the main text, we report results for five dialects that exhibited the lowest average performance across baseline models: Rural African American Vernacular English (RAAVE), Jamaican English, Aboriginal English, Welsh English, and Southwest England Dialect (SWE). Complete results for all evaluated dialects are provided in Appendix~\ref{sec:appendix}.


\vspace{2mm} \noindent \textbf{Evaluation Metrics.} We evaluate model performance using different metrics suited to each dataset. For \textbf{PrivacyQA}, we use the F1 score at the answer classification level. This metric is appropriate since PrivacyQA is framed as a sentence selection task, where models must determine whether a given sentence correctly answers a privacy-related question.

For \textbf{PolicyQA}, we adopt a token-level F1 score, commonly used in extractive question-answering tasks. This metric calculates the overlap between predicted answer spans and ground-truth answers at the token level. This approach ensures a fair assessment of partial matches, as PolicyQA requires extracting precise answer spans from privacy policy text rather than classifying entire sentences. We also compare the average difference between SAE and the other dialects and the maximum difference for both datasets.

\vspace{2mm} \noindent \textbf{Baselines.} We evaluate three models in this paper: Llama 3.1 8B~\cite{dubey2024llama}, DeepSeek-R1-Distill-Qwen-14B~\cite{guo2025deepseek}, and GPT-4o-mini~\cite{hurst2024gpt}. All models are evaluated in zero- and few-shot settings. Moreover, we evaluate them with our multi-agent framework with and without few-shot examples.

\vspace{2mm} \noindent \textbf{Results.} 
We evaluate our multi-agent framework on the PrivacyQA and PolicyQA datasets across SAE and five non-standard English dialects: Rural African American Vernacular English (RAAVE), Jamaican English, Aboriginal English, Welsh English, and Southwest England Dialect (SWE).

\begin{table}[t]
\small 
\centering

\resizebox{\linewidth}{!}{%
\begin{tabular}{lcccc}
\toprule
& \multicolumn{2}{c}{\textbf{PrivacyQA}} & \multicolumn{2}{c}{\textbf{PolicyQA}} \\
\cmidrule(lr){2-3}\cmidrule(lr){4-5}
\textbf{Setting} & \textbf{Initial ($\uparrow$)} & \textbf{Final ($\uparrow$)} & \textbf{Initial ($\uparrow$)} & \textbf{Final ($\uparrow$)} \\
\midrule
Zero-shot & .53 & \textbf{.59} & .43 & \textbf{.45} \\
Few-shot  & .58 & \textbf{.61} & .47 & \textbf{.48} \\
\bottomrule
\end{tabular}}
\caption{Ablation on \emph{Initial} vs.\ \emph{Final} answers for GPT-4o-mini before completing multiple back-and-forths between the Dialect and Policy Agents.  Scores are averaged across all English dialects.}
\label{tab:gpt4-small-ablation}
\end{table}

Table~\ref{tab:empty-table-1} presents the PrivacyQA results. Our multi-agent framework consistently improves performance across all dialects compared to baseline models. Notably, the GPT-4o-mini Multi-agent-few model achieves the highest average accuracy (0.598), outperforming its few-shot baseline (0.565). The average performance disparity (AVG Diff) is also reduced, with our multi-agent framework achieving a minimum AVG Diff of 0.005, compared to 0.016 in the best-performing baseline. \textcolor{black}{This reduction in disparity underscores the framework’s ability to generalize linguistic fairness across dialects, not just improve raw performance.} A similar trend is observed for Llama 3.1 and DeepSeek-R1, where our framework yields notable improvements. The Llama 3.1 Multi-agent-few improves overall performance to 0.530 while reducing AVG Diff to 0.008. DeepSeek-R1 Multi-agent-zero achieves the lowest Max Diff (0.017) among all models, indicating improved fairness across dialects.

\textcolor{black}{These improvements are not limited to low-resource dialects. We observe that even performance on SAE increases slightly in the multi-agent setup, suggesting that the collaborative refinement process benefits all users, not only those using non-standard varieties. Additionally, the DeepSeek-R1 Multi-agent-few model, while showing a slight drop in SAE, achieves substantial gains on challenging dialects like RAAVE (+.0967 over zero-shot) and Jamaican English (+.038), demonstrating the framework’s ability to reallocate capacity toward fairness without large performance trade-offs.}

\begin{figure}[t]
    \centering
    \includegraphics[width=\linewidth]{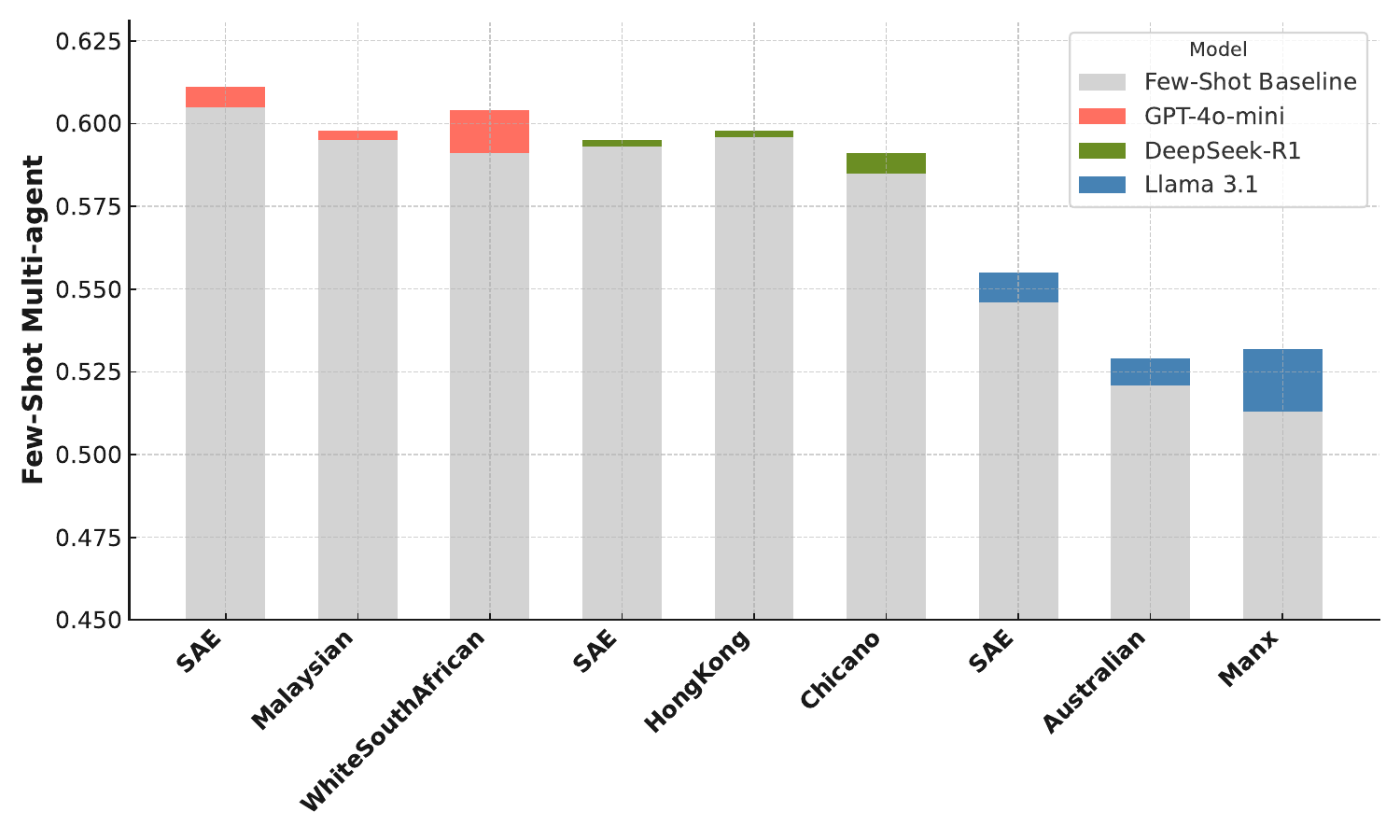}
    \caption{Comparison of the few-shot baseline performance (grey) $F_1$ scores with the improvements achieved by our method (colored bars) for each model on PrivacyQA. We compare SAE with the two highest-performing dialects for each model.} \vspace{-0em}
    \label{fig:topmodels}
\end{figure}

Table~\ref{tab:empty-table-2} shows results for PolicyQA. Our framework again enhances both overall performance and fairness. The GPT-4o-mini Multi-agent-few model achieves an average accuracy of 0.471, improving over the best baseline model (0.449). The disparity across dialects is also reduced, with our framework achieving an AVG Diff of 0.006, compared to 0.014 in the best baseline. For Llama 3.1, our framework improves overall accuracy from 0.371 (few-shot baseline) to 0.387 (multi-agent-few), reducing Max Diff from 0.080 to 0.028. Similarly, DeepSeek-R1 Multi-agent-few achieves an AVG Diff of 0.006, marking a substantial improvement in fairness.

\textcolor{black}{In contrast to PrivacyQA, where zero-shot models struggled more, PolicyQA exhibits overall tighter performance bands, making fairness improvements particularly notable. For example, the Llama 3.1 Multi-agent-few model reduces the Max Diff by more than half (from 0.080 to 0.028) while also achieving the highest gains on dialects such as Jamaican and Aboriginal English, with improvements of +.031 and +.028, respectively. DeepSeek-R1 similarly benefits, achieving a high average accuracy of 0.482 with one of the lowest disparities (AVG Diff = 0.006), which demonstrates that the benefits of our multi-agent design generalize across various question formats and task setups.}

One of the most striking findings is that the zero-shot performance of our multi-agent framework matches or even surpasses that of the few-shot baselines across multiple models. This demonstrates the ability of our approach to enhance performance without requiring additional in-context examples, making it highly effective in settings where labeled data is limited.

Across both datasets, our multi-agent framework substantially reduces performance disparities between SAE and non-standard dialects. Compared to baseline models, it consistently lowers Max Diff values, demonstrating improved fairness. At the same time, it improves absolute accuracy across all dialects, highlighting its effectiveness in mitigating dialectal biases in privacy-related QA systems.

\begin{table}[t] 
\centering
\footnotesize
\resizebox{.8\linewidth}{!}{%
\begin{tabular}{lcc}
\toprule
\textbf{Approach} & \textbf{Initial ($\uparrow$)} & \textbf{Final ($\uparrow$)} \\
\midrule
With Dialect Info & .5772 & .5966 \\
No Dialect Info & .5210 & .5894 \\
\bottomrule
\end{tabular}}
\caption{Average $F_1$ across dialects on PrivacyQA dataset, comparing \emph{With} vs.\ \emph{Without} dialect-specific background information.}\vspace{-0em}
\label{tab:no-dialect-ablation}
\end{table}

\vspace{2mm} \noindent \textbf{Ablations and Analysis.} In Table~\ref{tab:gpt4-small-ablation}, we present an ablation focused on the benefit of the iterative collaboration between the Dialect Agent and the Privacy Policy Agent for GPT-4o-mini. We compare system performance at the initial stage---where a translated query is passed to the Privacy Policy Agent for a single-pass answer---against the Final stage, where the Dialect Agent evaluates the initial answer and provides feedback for refinement. We observe consistent improvements in both PrivacyQA (from .53 to .59  F\textsubscript{1} in zero-shot and .58 to .61 in few-shot) and PolicyQA (.43 to .45 in zero-shot and .47 to .48 in few-shot).
These improvements underscore that a single-pass translation of dialectal queries does not fully capture users' linguistic nuances. While the dialect information helps a lot initially, once the Dialect Agent reviews the Privacy Policy Agent's answer, it corrects subtle misunderstandings (e.g., colloquial phrasing, dialect-specific grammatical structures), leading to more accurate final predictions. Notably, improvements persist in both zero-shot and few-shot settings, suggesting that agents' collaboration is effective even without additional in-context examples.

\begin{table}[t] 
\centering
\footnotesize
\resizebox{1.0\linewidth}{!}{%
\begin{tabular}{lcc}
\toprule
\textbf{Metric} & \textbf{Zero-shot} & \textbf{Few-shot} \\
\midrule
\textcolor{black}{Disagreements (Overrides)} & 22.99\% & 31.75\% \\
\textcolor{black}{Beneficial among Disagreements} & 63.4\% & 72.1\% \\
\textcolor{black}{Detrimental among Disagreements} & 24.1\% & 18.7\% \\
\bottomrule
\end{tabular}}
\caption{\textcolor{black}{Frequency and impact of Dialect Agent overrides on PrivacyQA}}
\label{tab:override-stats}
\end{table}

Figure~\ref{fig:topmodels} shows how our multi-agent framework improves performance compared to the few-shot baseline on PrivacyQA. The grey bars represent the few-shot baseline, while the colored bars show the improvements from our method.  We compare SAE for each model to the top two performing dialects on each model. Overall, we find that our approach improves the top-performing dialects as well. It does not only improve dialects the model does not perform well on (e.g., we see an improvement for SAE). We also find one interesting phenomenon, i.e., DeepSeek-R1 performs best on the Hong Kong English dialect, not SAE.

Next, we investigate the impact of removing dialect-specific background information (e.g., grammar and phonetic features) from the Dialect Agent’s prompt. Intuitively, we may not have access to or even know the dialectal information in complete detail. Hence, here we just prompt with ``You are a linguistics expert in English dialects,'' without even the dialect name. As shown in Table~\ref{tab:no-dialect-ablation}, omitting these linguistic details leads to performance declines at the \emph{Initial} stage (single-pass answer), dropping from 0.5772 to 0.5210 in average $F_1$. Although the \emph{Final} stage (after iterative refinement) still yields an improvement (up to 0.5894), the performance remains below that of the fully informed system, which reaches 0.5966. \textcolor{black}{Still, even without dialect metadata, the \emph{Final} stage model improves over the best-performing single-agent baseline (0.5602), yielding +2.9 F\textsubscript{1}}. This highlights that explicit knowledge of dialect-specific characteristics is critical for accurately interpreting user queries in non-standard English variants. Even with iterative agent collaboration, the absence of tailored dialect information constrains how effectively the system can capture nuanced morphological or syntactic cues, eventually reducing the correctness of privacy-policy answers. Please see Appendix~\ref{sec:erroranal} for a complete error analysis.

\textcolor{black}{Finally, we quantify how often the Dialect Agent intervenes and the effect of those interventions. As shown in Table~\ref{tab:override-stats}, the Dialect agent overrides the Privacy Policy Agent’s initial answer in 22.99\% of zero-shot cases and 31.75\% in few-shot cases. Among these overrides, 63.4\% are beneficial in the zero-shot setting (i.e., correcting an initial error), while 24.1\% are detrimental (i.e., introducing a new error). In few-shot, the success rate improves further, with 72.1\% of overrides helping and only 18.7\% hurting. These results suggest that the Dialect Agent plays a valuable corrective role, refining the output in most cases and contributing meaningfully to the overall performance improvements of the system.}

\textcolor{black}{We also observe that override rates vary across dialects, ranging from 14\% to 33\% (zero-shot) and 16\% to 43\% (few-shot). Roughly 9\% of overrides were neutral, where both initial and final responses were incorrect. These findings highlight the Dialect Agent’s consistent corrective role, particularly for dialects with greater divergence from SAE.
}

 \textcolor{black}{Finally, to assess the quality of these standardized translations (the final translation by the Dialect Agents), we compare them against the original human-authored references in the dataset. The translations achieve a BLEU score of 46.5 and a ROUGE-L score of 80.5, indicating that the Dialect Agent produces fluent and semantically faithful paraphrases of the original dialectal queries. Representative examples of these translations are provided in Appendix~\ref{appendix:translations}.}

\vspace{2mm} \noindent \textbf{Implications.} Our results highlight the critical role of incorporating dialect and cultural context in NLP systems. We demonstrate that even when no training data is available for a given dialect, providing minimal but targeted information about the dialect in the prompt can substantially improve model performance. This underscores the importance of designing NLP systems with a deep understanding of their potential users, ensuring that prompts account for linguistic and cultural variations. 

\textcolor{black}{Additionally, dialect-aware prompting strategies can serve as lightweight, scalable interventions for fairness in settings where large-scale data collection is infeasible or ethically complex, such as healthcare, legal reasoning, education, or multilingual customer service. In such domains, user trust and accessibility hinge on a system’s ability to reflect users’ linguistic identities.} 


\textcolor{black}{We acknowledge that explicit dialect labels may not always be available; future work should explore privacy-preserving, unsupervised methods to infer dialectal features directly from user queries.} Responsible AI development must extend beyond model selection and fine-tuning. Practitioners must carefully consider how their models interact with diverse user populations and adapt their prompting strategies accordingly. The success of our approach suggests that small, well-informed modifications to prompting strategies can have a meaningful impact, even in zero-shot settings. \textcolor{black}{ Looking ahead, we encourage future research on automated dialect detection, richer cultural representations in prompts, and end-to-end integration of multi-agent reasoning to build truly inclusive NLP systems.}

\section{Conclusion}
This work introduces a multi-agent framework to mitigate dialectal biases in privacy question-answering systems. Our approach reduces performance disparities across dialects while improving overall accuracy, demonstrating that incorporating dialect and cultural awareness can enhance NLP model fairness without requiring additional training data. By leveraging targeted prompts, our method achieves results comparable to or better than few-shot baselines in a zero-shot setting, underscoring the potential of structured prompting for equitable NLP applications.

These findings highlight the importance of accounting for linguistic diversity when designing NLP systems. Making language models accessible to users from diverse backgrounds requires prompting strategies that reflect dialectal variation. Future work should explore extending this approach to high-stakes domains such as healthcare, legal AI, and financial services, where language accessibility is critical. It is also important to investigate how dynamically adapting prompts based on user dialect can enhance real-time interactions with LLMs. Finally, exploring automated dialect detection mechanisms (e.g., in multicultural households) and integrating multi-agent collaboration into broader NLP pipelines could further advance fairness and inclusivity in large-scale language models.

\section*{Acknowledgments}
This material is based upon work supported by the National Science Foundation (NSF) under Grant~No. 2145357.

\section*{Limitations}
While our multi-agent framework effectively mitigates dialect biases in privacy policy QA, it has several limitations. First, our approach relies on synthetic dialectal data generated using rule-based transformations, which may not fully capture the nuances of naturally occurring dialect variations. Future work should evaluate performance on real-world dialectal data and user-generated queries to ensure robustness. Second, while our framework reduces performance disparities, some dialects still exhibit lower accuracy compared to Standard American English (SAE). This suggests that further refinements in the Dialect Agent's translation capabilities may be needed to preserve contextual nuances more effectively. Third, our method depends on accurate dialect metadata to select the appropriate linguistic adaptation strategy. In cases where dialect information is unavailable or ambiguous, performance gains may be limited. Finally, our study focuses on English dialects, and it remains an open question how well this framework generalizes to other languages with diverse linguistic variations.

\section*{Ethical Implications}
Our work highlights important ethical considerations in the development of NLP systems, particularly for high-stakes applications like privacy policy QA. By reducing dialectal disparities, our framework improves access to critical privacy information for speakers of non-standard English varieties, promoting fairness and inclusivity. However, dialect adaptation raises concerns about linguistic representation and cultural preservation. While translation into SAE may improve comprehension, it may also reinforce dominant linguistic norms at the expense of dialectal authenticity. Future research should explore methods that balance accessibility with dialectal preservation, ensuring that speakers of all linguistic backgrounds feel represented in NLP systems. Additionally, our study underscores the broader need for AI systems to consider sociolinguistic diversity in their design. Developers must be mindful of biases in training data, evaluation metrics, and system outputs to avoid perpetuating inequities in AI-driven decision-making. Further, our approach requires transparency in how dialect adaptation decisions are made, emphasizing the need for user agency in interacting with privacy policy QA systems.


\bibliography{custom}

@inproceedings{ziems2023multivalue,
 address = {Toronto, Canada},
 author = {Ziems, Caleb  and
Held, William  and
Yang, Jingfeng  and
Dhamala, Jwala  and
Gupta, Rahul  and
Yang, Diyi},
 booktitle = {Proceedings of the 61st Annual Meeting of the Association for Computational Linguistics (Volume 1: Long Papers)},
 doi = {10.18653/v1/2023.acl-long.44},
 editor = {Rogers, Anna  and
Boyd-Graber, Jordan  and
Okazaki, Naoaki},
 pages = {744--768},
 publisher = {Association for Computational Linguistics},
 title = {Multi-{VALUE}: A Framework for Cross-Dialectal {E}nglish {NLP}},
 url = {https://aclanthology.org/2023.acl-long.44},
 year = {2023}
}

@inproceedings{jorgensen2016learning,
 address = {San Diego, California},
 author = {J{\o}rgensen, Anna  and
Hovy, Dirk  and
S{\o}gaard, Anders},
 booktitle = {Proceedings of the 2016 Conference of the North {A}merican Chapter of the Association for Computational Linguistics: Human Language Technologies},
 doi = {10.18653/v1/N16-1130},
 editor = {Knight, Kevin  and
Nenkova, Ani  and
Rambow, Owen},
 pages = {1115--1120},
 publisher = {Association for Computational Linguistics},
 title = {Learning a {POS} tagger for {AAVE}-like language},
 url = {https://aclanthology.org/N16-1130},
 year = {2016}
}

@inproceedings{blodgett2018aaedisparities,
  title={Twitter universal dependency parsing for African-American and mainstream American English},
  author={Blodgett, Su Lin and Wei, Johnny and O’Connor, Brendan},
  booktitle={Proceedings of the 56th Annual Meeting of the Association for Computational Linguistics (Volume 1: Long Papers)},
  pages={1415--1425},
  year={2018}
}

@inproceedings{jurgens2017languageid,
 author = {Jurgens, David and Finethy, Tim and McCorriston, James and Xu, Yi Tian and Ruths, Derek},
 booktitle = {Proceedings of the Ninth International Conference on Web and Social Media},
 pages = {188--197},
 title = {Geolocation Prediction in Twitter Using Social Networks: A Critical Analysis and Review of Current Practice},
 year = {2017}
}

@inproceedings{sap2019socialbias,
 address = {Florence, Italy},
 author = {Sap, Maarten  and
Card, Dallas  and
Gabriel, Saadia  and
Choi, Yejin  and
Smith, Noah A.},
 booktitle = {Proceedings of the 57th Annual Meeting of the Association for Computational Linguistics},
 doi = {10.18653/v1/P19-1163},
 editor = {Korhonen, Anna  and
Traum, David  and
M{\`a}rquez, Llu{\'\i}s},
 pages = {1668--1678},
 publisher = {Association for Computational Linguistics},
 title = {The Risk of Racial Bias in Hate Speech Detection},
 url = {https://aclanthology.org/P19-1163},
 year = {2019}
}

@inproceedings{davidson2019hatebias,
 address = {Florence, Italy},
 author = {Davidson, Thomas  and
Bhattacharya, Debasmita  and
Weber, Ingmar},
 booktitle = {Proceedings of the Third Workshop on Abusive Language Online},
 doi = {10.18653/v1/W19-3504},
 editor = {Roberts, Sarah T.  and
Tetreault, Joel  and
Prabhakaran, Vinodkumar  and
Waseem, Zeerak},
 pages = {25--35},
 publisher = {Association for Computational Linguistics},
 title = {Racial Bias in Hate Speech and Abusive Language Detection Datasets},
 url = {https://aclanthology.org/W19-3504},
 year = {2019}
}

@inproceedings{liu2023dada,
 address = {Singapore},
 author = {Liu, Yanchen  and
Held, William  and
Yang, Diyi},
 booktitle = {Proceedings of the 2023 Conference on Empirical Methods in Natural Language Processing},
 doi = {10.18653/v1/2023.emnlp-main.850},
 editor = {Bouamor, Houda  and
Pino, Juan  and
Bali, Kalika},
 pages = {13776--13793},
 publisher = {Association for Computational Linguistics},
 title = {{DADA}: Dialect Adaptation via Dynamic Aggregation of Linguistic Rules},
 url = {https://aclanthology.org/2023.emnlp-main.850},
 year = {2023}
}

@inproceedings{held2023tada,
 address = {Toronto, Canada},
 author = {Held, William  and
Ziems, Caleb  and
Yang, Diyi},
 booktitle = {Findings of the Association for Computational Linguistics: ACL 2023},
 doi = {10.18653/v1/2023.findings-acl.51},
 editor = {Rogers, Anna  and
Boyd-Graber, Jordan  and
Okazaki, Naoaki},
 pages = {813--824},
 publisher = {Association for Computational Linguistics},
 title = {{TADA} : Task Agnostic Dialect Adapters for {E}nglish},
 url = {https://aclanthology.org/2023.findings-acl.51},
 year = {2023}
}

@article{kortmann2020ewave,
  title={Introducing the Platform of the Electronic World Atlas of Varieties of English (eWAVE).},
  author={Kortmann, Bernd and Lunkenheimer, Kerstin and Ehret, Katharina},
  journal={Research in English Language Pedagogy (RELP)},
  volume={13},
  number={4},
  year={2025}
}

@article{bender2021allocationalharms,
 author = {Bender, Emily M and Gebru, Timnit and McMillan-Major, Angelina and Shmitchell, Shmargaret},
 journal = {Proceedings of the 2021 ACM Conference on Fairness, Accountability, and Transparency},
 pages = {610--623},
 title = {On the Dangers of Stochastic Parrots: Can Language Models Be Too Big?},
 year = {2021}
}

@inproceedings{hovy2016nlpequity,
  title={The social impact of natural language processing},
  author={Hovy, Dirk and Spruit, Shannon L and others},
  booktitle={The 54th Annual Meeting of the Association for Computational Linguistics Proceedings of the Conference, Vol. 2 (Short Papers)},
  year={2016},
  organization={Association for Computational Linguistics}
}

@inproceedings{sun2023chinesedialects,
  title={Incorporating dialectal variability for socially equitable language identification},
  author={Jurgens, David and Tsvetkov, Yulia and Jurafsky, Dan},
  booktitle={Proceedings of the 55th Annual Meeting of the Association for Computational Linguistics (Volume 2: Short Papers)},
  pages={51--57},
  year={2017}
}

@inproceedings{chi2023plue,
 address = {Toronto, Canada},
 author = {Chi, Jianfeng  and
Ahmad, Wasi Uddin  and
Tian, Yuan  and
Chang, Kai-Wei},
 booktitle = {Proceedings of the 61st Annual Meeting of the Association for Computational Linguistics (Volume 2: Short Papers)},
 doi = {10.18653/v1/2023.acl-short.31},
 editor = {Rogers, Anna  and
Boyd-Graber, Jordan  and
Okazaki, Naoaki},
 pages = {352--365},
 publisher = {Association for Computational Linguistics},
 title = {{PLUE}: Language Understanding Evaluation Benchmark for Privacy Policies in {E}nglish},
 url = {https://aclanthology.org/2023.acl-short.31},
 year = {2023}
}

@inproceedings{ravichander2019question,
 address = {Hong Kong, China},
 author = {Ravichander, Abhilasha  and
Black, Alan W  and
Wilson, Shomir  and
Norton, Thomas  and
Sadeh, Norman},
 booktitle = {Proceedings of the 2019 Conference on Empirical Methods in Natural Language Processing and the 9th International Joint Conference on Natural Language Processing (EMNLP-IJCNLP)},
 doi = {10.18653/v1/D19-1500},
 editor = {Inui, Kentaro  and
Jiang, Jing  and
Ng, Vincent  and
Wan, Xiaojun},
 pages = {4947--4958},
 publisher = {Association for Computational Linguistics},
 title = {Question Answering for Privacy Policies: Combining Computational and Legal Perspectives},
 url = {https://aclanthology.org/D19-1500},
 year = {2019}
}

@misc{EPIC_Privacy_Racial_Justice,
  author = "{Electronic Privacy Information Center}",
  title = "{Privacy and Racial Justice}",
  year = "2025",
  url = "https://epic.org/issues/democracy-free-speech/privacy-and-racial-justice/",
  note = "Accessed: 2025-02-16"
}

@inproceedings{ahmad2020policyqa,
 address = {Online},
 author = {Ahmad, Wasi  and
Chi, Jianfeng  and
Tian, Yuan  and
Chang, Kai-Wei},
 booktitle = {Findings of the Association for Computational Linguistics: EMNLP 2020},
 doi = {10.18653/v1/2020.findings-emnlp.66},
 editor = {Cohn, Trevor  and
He, Yulan  and
Liu, Yang},
 pages = {743--749},
 publisher = {Association for Computational Linguistics},
 title = {{P}olicy{QA}: A Reading Comprehension Dataset for Privacy Policies},
 url = {https://aclanthology.org/2020.findings-emnlp.66},
 year = {2020}
}

@inproceedings{ahmad2021policyie,
 address = {Online},
 author = {Ahmad, Wasi  and
Chi, Jianfeng  and
Le, Tu  and
Norton, Thomas  and
Tian, Yuan  and
Chang, Kai-Wei},
 booktitle = {Proceedings of the 59th Annual Meeting of the Association for Computational Linguistics and the 11th International Joint Conference on Natural Language Processing (Volume 1: Long Papers)},
 doi = {10.18653/v1/2021.acl-long.340},
 editor = {Zong, Chengqing  and
Xia, Fei  and
Li, Wenjie  and
Navigli, Roberto},
 pages = {4402--4417},
 publisher = {Association for Computational Linguistics},
 title = {Intent Classification and Slot Filling for Privacy Policies},
 url = {https://aclanthology.org/2021.acl-long.340},
 year = {2021}
}

@article{bui2021piextract,
  title={Automated extraction and presentation of data practices in privacy policies},
  author={Bui, Duc and Shin, Kang G and Choi, Jong-Min and Shin, Junbum},
  journal={Proceedings on Privacy Enhancing Technologies},
  year={2021}
}

@inproceedings{zhao2024longagent,
 author = {Jun Zhao and Can Zu and Hao Xu and Yi Lu and Wei He and Yiwen Ding and Tao Gui and Qi Zhang and Xuanjing Huang},
 booktitle = {Proceedings of the 2024 Conference on Empirical Methods in Natural Language Processing},
 pages = {16310--16324},
 title = {LONGAGENT: Achieving Question Answering for 128k-Token-Long Documents through Multi-Agent Collaboration},
 year = {2024}
}

@inproceedings{zhao2024gedi,
 author = {Xiutian Zhao and Ke Wang and Wei Peng},
 booktitle = {Proceedings of the 2024 Conference on Empirical Methods in Natural Language Processing},
 pages = {2712--2727},
 title = {An Electoral Approach to Diversify LLM-based Multi-Agent Collective Decision-Making},
 year = {2024}
}

@article{chen2023multiagentreasoning,
  title={Tdag: A multi-agent framework based on dynamic task decomposition and agent generation},
  author={Wang, Yaoxiang and Wu, Zhiyong and Yao, Junfeng and Su, Jinsong},
  journal={Neural Networks},
  volume={185},
  pages={107200},
  year={2025},
  publisher={Elsevier}
}

@article{lwowski2021risk,
 author = {Lwowski, B and Rios, A},
 journal = {Journal of the American Medical Informatics Association: JAMIA},
 number = {4},
 pages = {839--849},
 title = {The risk of racial bias while tracking influenza-related content on social media using machine learning.},
 volume = {28},
 year = {2021}
}

@inproceedings{lee2022type,
 address = {Gyeongju, Republic of Korea},
 author = {Lee, Seungyeon  and
Lee, Minho},
 booktitle = {Proceedings of the 29th International Conference on Computational Linguistics},
 editor = {Calzolari, Nicoletta  and
Huang, Chu-Ren  and
Kim, Hansaem  and
Pustejovsky, James  and
Wanner, Leo  and
Choi, Key-Sun  and
Ryu, Pum-Mo  and
Chen, Hsin-Hsi  and
Donatelli, Lucia  and
Ji, Heng  and
Kurohashi, Sadao  and
Paggio, Patrizia  and
Xue, Nianwen  and
Kim, Seokhwan  and
Hahm, Younggyun  and
He, Zhong  and
Lee, Tony Kyungil  and
Santus, Enrico  and
Bond, Francis  and
Na, Seung-Hoon},
 pages = {6301--6314},
 publisher = {International Committee on Computational Linguistics},
 title = {Type-dependent Prompt {C}ycle{QAG} : Cycle Consistency for Multi-hop Question Generation},
 url = {https://aclanthology.org/2022.coling-1.549},
 year = {2022}
}

@inproceedings{yu2023exploring,
 address = {Toronto, Canada},
 author = {Yu, Fangyi  and
Quartey, Lee  and
Schilder, Frank},
 booktitle = {Findings of the Association for Computational Linguistics: ACL 2023},
 doi = {10.18653/v1/2023.findings-acl.858},
 editor = {Rogers, Anna  and
Boyd-Graber, Jordan  and
Okazaki, Naoaki},
 pages = {13582--13596},
 publisher = {Association for Computational Linguistics},
 title = {Exploring the Effectiveness of Prompt Engineering for Legal Reasoning Tasks},
 url = {https://aclanthology.org/2023.findings-acl.858},
 year = {2023}
}

@inproceedings{faisal-etal-2024-dialectbench,
 address = {Bangkok, Thailand},
 author = {Faisal, Fahim  and
Ahia, Orevaoghene  and
Srivastava, Aarohi  and
Ahuja, Kabir  and
Chiang, David  and
Tsvetkov, Yulia  and
Anastasopoulos, Antonios},
 booktitle = {Proceedings of the 62nd Annual Meeting of the Association for Computational Linguistics (Volume 1: Long Papers)},
 doi = {10.18653/v1/2024.acl-long.777},
 editor = {Ku, Lun-Wei  and
Martins, Andre  and
Srikumar, Vivek},
 pages = {14412--14454},
 publisher = {Association for Computational Linguistics},
 title = {{DIALECTBENCH}: An {NLP} Benchmark for Dialects, Varieties, and Closely-Related Languages},
 url = {https://aclanthology.org/2024.acl-long.777/},
 year = {2024}
}

@inproceedings{naous-etal-2024-beer,
 address = {Bangkok, Thailand},
 author = {Naous, Tarek  and
Ryan, Michael J  and
Ritter, Alan  and
Xu, Wei},
 booktitle = {Proceedings of the 62nd Annual Meeting of the Association for Computational Linguistics (Volume 1: Long Papers)},
 doi = {10.18653/v1/2024.acl-long.862},
 editor = {Ku, Lun-Wei  and
Martins, Andre  and
Srikumar, Vivek},
 pages = {16366--16393},
 publisher = {Association for Computational Linguistics},
 title = {Having Beer after Prayer? Measuring Cultural Bias in Large Language Models},
 url = {https://aclanthology.org/2024.acl-long.862/},
 year = {2024}
}

@article{liu2024culturally,
 author = {Liu, Chen Cecilia and Gurevych, Iryna and Korhonen, Anna},
 journal = {ArXiv preprint},
 title = {Culturally Aware and Adapted NLP: A Taxonomy and a Survey of the State of the Art},
 url = {https://arxiv.org/abs/2406.03930},
 volume = {abs/2406.03930},
 year = {2024}
}

@inproceedings{lwowski2022measuring,
 address = {Gyeongju, Republic of Korea},
 author = {Lwowski, Brandon  and
Rad, Paul  and
Rios, Anthony},
 booktitle = {Proceedings of the 29th International Conference on Computational Linguistics},
 editor = {Calzolari, Nicoletta  and
Huang, Chu-Ren  and
Kim, Hansaem  and
Pustejovsky, James  and
Wanner, Leo  and
Choi, Key-Sun  and
Ryu, Pum-Mo  and
Chen, Hsin-Hsi  and
Donatelli, Lucia  and
Ji, Heng  and
Kurohashi, Sadao  and
Paggio, Patrizia  and
Xue, Nianwen  and
Kim, Seokhwan  and
Hahm, Younggyun  and
He, Zhong  and
Lee, Tony Kyungil  and
Santus, Enrico  and
Bond, Francis  and
Na, Seung-Hoon},
 pages = {6600--6616},
 publisher = {International Committee on Computational Linguistics},
 title = {Measuring Geographic Performance Disparities of Offensive Language Classifiers},
 url = {https://aclanthology.org/2022.coling-1.574},
 year = {2022}
}

@article{hurst2024gpt,
  title={Gpt-4o system card},
  author={Hurst, Aaron and Lerer, Adam and Goucher, Adam P and Perelman, Adam and Ramesh, Aditya and Clark, Aidan and Ostrow, AJ and Welihinda, Akila and Hayes, Alan and Radford, Alec and others},
  journal={arXiv preprint arXiv:2410.21276},
  year={2024}
}

@article{guo2025deepseek,
  title={Deepseek-r1: Incentivizing reasoning capability in llms via reinforcement learning},
  author={Guo, Daya and Yang, Dejian and Zhang, Haowei and Song, Junxiao and Zhang, Ruoyu and Xu, Runxin and Zhu, Qihao and Ma, Shirong and Wang, Peiyi and Bi, Xiao and others},
  journal={arXiv preprint arXiv:2501.12948},
  year={2025}
}

@article{dubey2024llama,
  title={The llama 3 herd of models},
  author={Dubey, Abhimanyu and Jauhri, Abhinav and Pandey, Abhinav and Kadian, Abhishek and Al-Dahle, Ahmad and Letman, Aiesha and Mathur, Akhil and Schelten, Alan and Yang, Amy and Fan, Angela and others},
  journal={arXiv preprint arXiv:2407.21783},
  year={2024}
}

@article{cooley2000human,
  title={Human-centered design},
  author={Cooley, Mike},
  journal={Information design},
  pages={59--81},
  year={2000},
  publisher={MIT press}
}

@inproceedings{liu2023knowledgeagents,
  title={Hm-rag: Hierarchical multi-agent multimodal retrieval augmented generation},
  author={Liu, Pei and Liu, Xin and Yao, Ruoyu and Liu, Junming and Meng, Siyuan and Wang, Ding and Ma, Jun},
  booktitle={Proceedings of the 33rd ACM international conference on multimedia},
  pages={2781--2790},
  year={2025}
}

@article{xu2023structureddecision,
  title={A comprehensive survey on multi-agent cooperative decision-making: Scenarios, approaches, challenges and perspectives},
  author={Jin, Weiqiang and Du, Hongyang and Zhao, Biao and Tian, Xingwu and Shi, Bohang and Yang, Guang},
  journal={arXiv preprint arXiv:2503.13415},
  year={2025}
}

@inproceedings{wilson2016creation,
  title={The creation and analysis of a website privacy policy corpus},
  author={Wilson, S. and others},
  booktitle={Proceedings of the 54th Annual Meeting of the Association for Computational Linguistics (Volume 1: Long Papers)},
  pages={1330--1340},
  year={2016}
}

@inproceedings{srinath2021privacy,
  title={Privacy at Scale: Introducing the PrivaSeer Corpus of Web Privacy Policies},
  author={Srinath, M. and Wilson, S. and Giles, C. L.},
  booktitle={Proceedings of the 59th Annual Meeting of the Association for Computational Linguistics and the 11th International Joint Conference on Natural Language Processing},
  pages={6829--6839},
  year={2021}
}

@inproceedings{manandhar2022smart,
  title={Smart Home Privacy Policies Demystified: A Study of Availability, Content, and Coverage},
  author={Manandhar, S. and others},
  booktitle={31st USENIX Security Symposium},
  pages={3521--3538},
  year={2022}
}

@inproceedings{ramanath2014unsupervised,
  title={Unsupervised alignment of privacy policies using hidden Markov models},
  author={Ramanath, R. and Liu, F. and Sadeh, N. and Smith, N. A.},
  booktitle={Proceedings of ACL},
  year={2014}
}

@inproceedings{amos2021privacy,
  title={Privacy policies over time: Curation and analysis of a million-document dataset},
  author={Amos, R. and others},
  booktitle={Proceedings of the Web Conference},
  pages={2165--2176},
  year={2021}
}

@article{klisura,
  title={Unmasking database vulnerabilities: Zero-Knowledge schema inference attacks in Text-to-SQL systems},
  author={Klisura, {\DJ}or{\dj}e and Rios, Anthony},
  journal={arXiv preprint arXiv:2406.14545},
  year={2024}
}

@article{li2023camel,
  title={Camel: Communicative agents for" mind" exploration of large language model society},
  author={Li, Guohao and Hammoud, Hasan and Itani, Hani and Khizbullin, Dmitrii and Ghanem, Bernard},
  journal={Advances in Neural Information Processing Systems},
  volume={36},
  pages={51991--52008},
  year={2023}
}

@misc{chen2024autoagentsframeworkautomaticagent,
      title={AutoAgents: A Framework for Automatic Agent Generation}, 
      author={Guangyao Chen and Siwei Dong and Yu Shu and Ge Zhang and Jaward Sesay and Börje F. Karlsson and Jie Fu and Yemin Shi},
      year={2024},
      eprint={2309.17288},
      archivePrefix={arXiv},
      primaryClass={cs.AI},
      url={https://arxiv.org/abs/2309.17288}, 
}

@article{qian2023chatdev,
  title={Chatdev: Communicative agents for software development},
  author={Qian, Chen and Liu, Wei and Liu, Hongzhang and Chen, Nuo and Dang, Yufan and Li, Jiahao and Yang, Cheng and Chen, Weize and Su, Yusheng and Cong, Xin and others},
  journal={arXiv preprint arXiv:2307.07924},
  year={2023}
}

@article{li2024survey,
  title={A survey on LLM-based multi-agent systems: workflow, infrastructure, and challenges},
  author={Li, Xinyi and Wang, Sai and Zeng, Siqi and Wu, Yu and Yang, Yi},
  journal={Vicinagearth},
  volume={1},
  number={1},
  pages={9},
  year={2024},
  publisher={Springer}
}

@inproceedings{demszky2019analyzing,
  title={Analyzing polarization in social media: Method and application to tweets on 21 mass shootings},
  author={Demszky, Dorottya and Garg, Nikhil and Voigt, Rob and Zou, James and Shapiro, Jesse and Gentzkow, Matthew and Jurafsky, Dan},
  booktitle={Proceedings of the 2019 conference of the north american chapter of the association for computational linguistics: Human language technologies, volume 1 (long and short papers)},
  pages={2970--3005},
  year={2019}
}

\appendix
\section{Error Analysis}\label{sec:erroranal}
\vspace{2mm} \noindent Our error analysis indicates that performance variations across dialects likely stem from training data biases, as less-represented dialects consistently yielded lower final F1 scores, suggesting challenges in capturing subtle linguistic nuances. In some cases, the multiagent framework's refinement process yielded marginal improvements, yet in other examples, adjustments introduced new errors, particularly for dialects with complex or idiomatic expressions. 

In the PolicyQA task, for instance, one error involved the segment 
\begin{quote}
Last Updated on May 22, 2015    
\end{quote}
 paired with the question ``Do you take the user's opinion before or after making changes in policy?`` where the annotated answer was ``Last Updated on May 22, 2015``. This example shows how the model mistakenly extracted meta-information as the answer rather than identifying the procedural detail requested by the question. In another example, the question ``Does the privacy policy mention anything about children?`` was paired with a lengthy segment 
\begin{quote}
    You can jump to specific areas of our Privacy Policy by clicking on the links below, or you can read on for the full Privacy Policy: Information We Collect How We Use Personal Information We Collect How We May Disclose Personal Information We Collect How We May Use or Disclose Other Information We Collect Your Options How We Protect Your Personal Information Cookies Social Networking and Third Party Sites California Users' Privacy Rights Children's Online Privacy International Contact Us
\end{quote}

and the annotated answer was ``Children's.`` Here, the generative models' tendency to provide longer, more contextually diffuse answers led them to miss the succinct, targeted answer. These examples underscore a common issue with large language models: their inclination to generate overly verbose responses, which highlights the need for more targeted fine-tuning and improved context disambiguation for precise answer extraction.

\section{Prompts for Dialect and Privacy Policy Agents}
\label{appendix:agent-prompts}

To implement our multi-agent framework, we designed two specialized agents: the \textit{Dialect Agent} and the \textit{Privacy Policy Agent}. The \textit{Dialect Agent} is responsible for translating user queries from a given dialect into Standard American English (SAE) while preserving the original intent. Additionally, it plays a critical role in validating the responses generated by the Privacy Policy Agent. The \textit{Privacy Policy Agent} processes the translated queries, retrieving relevant information from a given privacy policy and determining whether a policy segment is \textit{Relevant} or \textit{Irrelevant} to the question.

The following subsections describe the prompts used to guide each agent at different stages of our method.

\subsection{Dialect Agent Prompts}
\label{sec:dialect-agent-prompts}

\subsubsection{Initial Translation Prompt}
The Dialect Agent first translates a user's query from a non-standard English dialect into Standard American English (SAE). This translation ensures that downstream processing by the Privacy Policy Agent is not negatively impacted by dialectal variations.

\begin{tcolorbox}[colback=gray!5!white, colframe=black, fontupper=\small, width=\linewidth, boxsep=0pt, title=Dialect Agent: Initial Translation]
\textbf{SYSTEM PROMPT}  

You are an expert linguist specializing in the following dialect:

\texttt{\{dialect\_info\}}

Your task is to translate the following question from this dialect into clear, Standard American English. Ensure that the translation is easily understandable to a general audience. Please provide only the translated question and do not include any additional text.

\textbf{USER MESSAGE}  

\texttt{\{question\}}
\end{tcolorbox}

At this stage, no feedback from the Privacy Policy Agent is available. The Dialect Agent simply returns the translated question.

\subsubsection{Responding to Expert Feedback}
After the Privacy Policy Agent classifies a privacy policy segment as \textit{Relevant} or \textit{Irrelevant}, the Dialect Agent evaluates whether the classification is consistent with the original intent of the user’s question in their dialect.

\begin{tcolorbox}[colback=gray!5!white, colframe=black, fontupper=\small, width=\linewidth, boxsep=0pt, title=Dialect Agent: Evaluating Privacy Agent’s Response]
\textbf{SYSTEM PROMPT}  

You are an expert linguist specializing in the following dialect, with expertise in privacy policies.

Previously, you translated a question from this dialect into Standard American English. Now, you need to critically assess whether the Privacy Policy Agent’s classification accurately reflects the meaning of the original question in the dialect.

\textbf{Privacy Policy Segment:}  

\texttt{\{privacy\_policy\_segment\}}

\textbf{Original Question in Dialect:}  

\texttt{\{question\}}

The Privacy Policy Agent has classified the policy segment as '\texttt{\{classification\}}' with the following reasoning:

\texttt{\{reasoning\}}

Based on your understanding of the dialect and its nuances, analyze the expert's classification and reasoning. Do you find any discrepancies or misunderstandings? Please provide a detailed explanation and conclude with either 'Agree' if you concur with the classification or 'Disagree' if you do not.
\end{tcolorbox}

If the Dialect Agent disagrees, the Privacy Policy Agent will be prompted to reconsider its classification based on the Dialect Agent’s insights.

\subsection{Privacy Policy Agent Prompts}
\label{sec:privacy-agent-prompts}

\subsubsection{Initial Classification Prompt}
The Privacy Policy Agent is responsible for determining whether a privacy policy segment is relevant to a user's question. In PrivacyQA, this classification is binary (\textit{Relevant} or \textit{Irrelevant}), while in PolicyQA, the Privacy Policy Agent provides a direct answer based on the policy text.

\begin{tcolorbox}[colback=gray!5!white, colframe=black, fontupper=\small, width=\linewidth, boxsep=0pt, title=Privacy Policy Agent: Initial Classification]
\textbf{SYSTEM PROMPT}  

You are a privacy policy expert. Your task is to determine whether the provided privacy policy segment is 'Relevant' or 'Irrelevant' to the question, based on the following definitions:

\textbf{Definitions:}  

- \textbf{Relevant}: The policy segment directly addresses the question.  

- \textbf{Irrelevant}: The policy segment does not directly address the question.

Please analyze the material below and provide:
1. A brief explanation of your reasoning.
2. Conclude only with 'Label: Relevant' or 'Label: Irrelevant'.

\textbf{USER MESSAGE}  

\textbf{Privacy Policy Segment:}  

\texttt{\{privacy\_policy\_segment\}}

\textbf{Question:}  

\texttt{\{translated\_question\}}
\end{tcolorbox}

In this zero-shot setup, the Privacy Policy Agent classifies the segment and explains its decision.

\subsubsection{Reconsideration Prompt (After Dialect Feedback)}

\begin{tcolorbox}[colback=gray!5!white, colframe=black, fontupper=\small, width=\linewidth, boxsep=0pt, title=Privacy Policy Agent: Reconsideration After Dialect Feedback]
\textbf{SYSTEM PROMPT}  

You are a privacy policy expert. Previously, you classified the privacy policy segment as '\texttt{\{previous\_classification\}}' regarding the question, with the following reasoning:

\texttt{\{previous\_reasoning\}}

However, the Dialect Agent has provided additional insights and \textbf{disagrees} with your classification. Their reasoning is as follows:

\texttt{\{dialect\_reasoning\}}

Please reconsider your initial decision in light of this new information. Provide:
1. A brief explanation of your reconsidered decision.
2. Conclude with 'Final Label: Relevant' or 'Final Label: Irrelevant'.
\end{tcolorbox}


\section{Dialect Details}\label{sec:dialect}

In this section, we provide examples of the dialect information we give to the LLMs to help them better understand linguistic variations. Each dialect entry includes key phonetic, grammatical, and vocabulary differences compared to Standard American English (SAE), along with cultural context. This information helps the model accurately translate dialectal queries while preserving their meaning. For example, Indian English includes retroflex consonants and distinct grammatical patterns, while Jamaican English (Patois) features non-rhotic pronunciation and unique verb structures. By incorporating these details, our framework improves the model’s ability to handle dialect-specific nuances in privacy policy question-answering.

Here is an example of the Indian English prompt:
\begin{center}

\resizebox{0.98\linewidth}{!}{%
\begin{tcolorbox}[colback=gray!5!white, colframe=black, fontupper=\small, width=\linewidth, boxsep=0pt, title=Indian Dialect]
    Key Features of Indian English\\
    
    Phonetics and Pronunciation:\\
    - Retroflex consonants influenced by Indian languages.\\
    - Variable stress and intonation patterns.\\
    - Vowel pronunciation often closer to native Indian languages.\\
    
    Grammar:\\
    - Use of present continuous for habitual actions (e.g., 'I am knowing').\\
    - Omission of articles and prepositions in certain contexts.\\
    - Use of Indian syntax and sentence structures.\\
    
    Vocabulary:\\
    - Incorporation of Hindi, Tamil, Bengali, and other Indian language terms\\
    - Unique expressions and idioms specific to Indian culture.\\
    
    Cultural Notes:\\
    - Reflects India's diverse linguistic landscape.\\
    - Widely used in Indian media, education, and business. \end{tcolorbox}}
\end{center}

Here is an example of the Jamaican English prompt:
\begin{center}
\resizebox{\linewidth}{!}{%
\begin{tcolorbox}[colback=gray!5!white, colframe=black, fontupper=\small, width=\linewidth, boxsep=0pt, title=Jamaican English] 
       Key Features of Jamaican English (Jamaican Patois)\\
    
    Phonetics and Pronunciation:\\
    - Non-rhotic pronunciation with 'r' often not pronounced.\\
    - Use of tone and pitch influenced by African languages.\\
    - Simplified consonant clusters and vowel shifts.\\
    
    Grammar:\\
    - Use of particles like 'fi' (to) and 'a' (progressive aspect).\\
    - Simplified tense markers and verb forms.\\
    - Use of double negatives for emphasis.\\
    
    Vocabulary:
    - Extensive borrowing from West African languages, Spanish, and English.\\
    - Unique slang and expressions reflecting Jamaican culture.\\
    
    Cultural Notes:
    - Central to Jamaican music genres like reggae and dancehall.\\
    - Reflects the island's history and multicultural influences.
\end{tcolorbox}} 
\end{center}

\begin{table*}[ht] 
\centering
\normalsize
\resizebox{1.0\linewidth}{!}{%
\begin{tabular}{p{0.48\linewidth} p{0.48\linewidth}}
\toprule
\textbf{Dialectal Input (AAVE)} & \textbf{Dialect Agent Translation (SAE)} \\
\midrule
It is access to my information? & Who is going to have access to my information? \\
gon for me test results be shared with any third party? & Will my test results be shared with any third-party? \\
what information it is access to that collaborators ? & What information do the collaborators have access to? \\
which information, if any, do that app sell to other people? & What information, if any, does that app sell to others? \\
do the app need any special permission for to run ? & Does the app need any special permissions to run? \\
\bottomrule
\end{tabular}}
\caption{Examples of AAVE queries and their SAE translations produced by the Dialect Agent. No hallucinated content was observed across over 500 spot-checked samples.}
\label{tab:translation-examples}
\end{table*}

\section{\textcolor{black}{Dialect Translation Examples}}
\label{appendix:translations}

\textcolor{black}{To evaluate the reliability of the Dialect Agent’s output, we manually inspected over 500 SAE translations produced by the agent when translating dialectal queries (e.g., AAVE) from the Multi-VALUE benchmark. We found no instances of hallucination, i.e., the agent did not invent new content, facts, or answer components. This outcome is expected given the bounded task design: translating sentence-level questions from dialectal English into Standard American English (SAE), often involving paraphrasing rather than generation from scratch.}

Table~\ref{tab:translation-examples} shows representative examples of AAVE queries and the Dialect Agent’s SAE translations. These illustrate how the agent improves clarity while maintaining user intent and factual fidelity.

\section{Resources}
All experiments were trained on a server with two NVIDIA A6000 GPUs.

\section{Full Results}
\label{sec:appendix}

This section shows all of the results for all 50 dialects generated using the Multi-Value framework. See Tables~\ref{tab:t1}, \ref{tab:t2}, \ref{tab:t3}, and \ref{tab:t4}. Table~\ref{tab:nodialectfull} shows the full dialect results without any specific dialect (they are a general dialect expert) information is passed directly to the dialect agent.

\begin{table*}[t]
\centering
\normalsize
\caption{Baseline Results for GPT-4, Llama 3.1, and DeepSeek-R1 on PrivacyQA (PQA) and PolicyQA (PoQA). 
``PQA 0'' = PrivacyQA Zero-shot, ``PQA F'' = PrivacyQA Few-shot, ``PoQA 0'' = PolicyQA Zero-shot, ``PoQA F'' = PolicyQA Few-shot.}\label{tab:t1}
\begin{adjustbox}{max width=\textwidth}
\begin{tabular}{lcccccccccccc}
\toprule
\textbf{Dialect} & \multicolumn{4}{c}{\textbf{GPT-4}} & \multicolumn{4}{c}{\textbf{Llama 3.1}} & \multicolumn{4}{c}{\textbf{DeepSeek-R1}}\\
& \textbf{PQA 0} & \textbf{PQA F} & \textbf{PoQA 0} & \textbf{PoQA F} & \textbf{PQA 0} & \textbf{PQA F} & \textbf{PoQA 0} & \textbf{PoQA F} & \textbf{PQA 0} & \textbf{PQA F} & \textbf{PoQA 0} & \textbf{PoQA F} \\
\midrule
Standard American Dialect & .394 & .605 & .352 & .478 & .469 & .546 & .310 & .412 & .532 & .581 & .455 & .446 \\
Kenyan Dialect & .386 & .595 & .337 & .439 & .430 & .465 & .247 & .380 & .536 & .570 & .425 & .466 \\
Sri Lankan Dialect & .386 & .595 & .336 & .447 & .438 & .453 & .256 & .371 & .531 & .571 & .435 & .500 \\
Scottish Dialect & .385 & .594 & .315 & .454 & .420 & .473 & .285 & .375 & .539 & .585 & .439 & .487 \\
Malaysian Dialect & .380 & .592 & .333 & .451 & .403 & .488 & .239 & .364 & .532 & .567 & .421 & .486 \\
Indian Dialect & .379 & .591 & .333 & .433 & .376 & .487 & .208 & .340 & .535 & .557 & .408 & .473 \\
Chicano Dialect & .379 & .580 & .320 & .441 & .456 & .467 & .287 & .365 & .532 & .591 & .458 & .498 \\
Cameroon Dialect & .378 & .580 & .342 & .430 & .390 & .484 & .246 & .348 & .541 & .539 & .453 & .475 \\
Ghanaian Dialect & .377 & .584 & .329 & .451 & .400 & .510 & .248 & .353 & .535 & .551 & .437 & .468 \\
Nigerian Dialect & .375 & .582 & .324 & .463 & .469 & .487 & .240 & .375 & .540 & .592 & .426 & .478 \\
Appalachian Dialect & .375 & .583 & .320 & .436 & .439 & .462 & .244 & .365 & .538 & .560 & .458 & .487 \\
White South African Dialect & .373 & .584 & .320 & .439 & .423 & .487 & .257 & .386 & .551 & .557 & .444 & .461 \\
Channel Islands Dialect & .372 & .581 & .324 & .438 & .431 & .465 & .263 & .376 & .538 & .559 & .409 & .456 \\
Southeast American Enclave Dialect & .372 & .579 & .328 & .444 & .391 & .370 & .262 & .370 & .551 & .563 & .432 & .475 \\
Ugandan Dialect & .372 & .578 & .331 & .449 & .433 & .470 & .246 & .363 & .551 & .578 & .422 & .453 \\
Liberian Settler Dialect & .371 & .577 & .326 & .444 & .377 & .478 & .270 & .376 & .553 & .545 & .417 & .481 \\
Cape Flats Dialect & .370 & .576 & .328 & .444 & .440 & .465 & .257 & .381 & .535 & .570 & .411 & .468 \\
Tristan Dialect & .368 & .575 & .324 & .439 & .393 & .466 & .251 & .339 & .540 & .549 & .447 & .466 \\
Ozark Dialect & .368 & .574 & .328 & .442 & .410 & .502 & .290 & .381 & .530 & .559 & .453 & .446 \\
Australian Dialect & .367 & .574 & .321 & .434 & .436 & .521 & .250 & .353 & .543 & .557 & .416 & .461 \\
Tanzanian Dialect & .366 & .573 & .333 & .452 & .401 & .482 & .264 & .382 & .536 & .569 & .446 & .509 \\
Fiji Acrolect & .364 & .572 & .333 & .445 & .409 & .500 & .265 & .382 & .557 & .570 & .458 & .475 \\
Fiji Basilect & .364 & .571 & .338 & .460 & .344 & .506 & .228 & .381 & .547 & .518 & .448 & .441 \\
Pakistani Dialect & .364 & .569 & .319 & .430 & .392 & .427 & .260 & .359 & .533 & .574 & .428 & .447 \\
Philippine Dialect & .363 & .568 & .349 & .471 & .370 & .506 & .240 & .366 & .552 & .548 & .440 & .479 \\
White Zimbabwean Dialect & .363 & .567 & .330 & .449 & .425 & .465 & .260 & .352 & .537 & .582 & .433 & .468 \\
Newfoundland Dialect & .362 & .566 & .319 & .428 & .394 & .508 & .264 & .374 & .526 & .556 & .420 & .493 \\
Orkney Shetland Dialect & .362 & .565 & .335 & .454 & .452 & .494 & .250 & .380 & .530 & .561 & .443 & .490 \\
East Anglican Dialect & .361 & .564 & .319 & .422 & .412 & .466 & .246 & .374 & .527 & .559 & .422 & .478 \\
Early African American Vernacular & .358 & .563 & .319 & .423 & .393 & .465 & .231 & .373 & .549 & .560 & .430 & .478 \\
Falkland Islands Dialect & .358 & .562 & .333 & .451 & .439 & .475 & .268 & .365 & .535 & .574 & .453 & .484 \\
Australian Vernacular & .357 & .561 & .329 & .448 & .398 & .479 & .240 & .387 & .537 & .579 & .453 & .468 \\
Black South African Dialect & .356 & .560 & .311 & .420 & .381 & .461 & .228 & .364 & .541 & .551 & .455 & .497 \\
Colloquial American Dialect & .354 & .559 & .326 & .443 & .375 & .489 & .276 & .361 & .526 & .572 & .439 & .471 \\
Indian South African Dialect & .353 & .558 & .336 & .454 & .377 & .467 & .207 & .352 & .541 & .554 & .447 & .459 \\
New Zealand Dialect & .353 & .557 & .344 & .464 & .387 & .494 & .241 & .345 & .550 & .567 & .434 & .473 \\
Bahamian Dialect & .352 & .556 & .325 & .441 & .345 & .458 & .241 & .352 & .537 & .526 & .448 & .473 \\
Hong Kong Dialect & .351 & .555 & .336 & .455 & .406 & .503 & .237 & .342 & .566 & .596 & .465 & .497 \\
Colloquial Singapore Dialect & .350 & .554 & .346 & .464 & .384 & .463 & .210 & .370 & .538 & .529 & .434 & .434 \\
Manx Dialect & .349 & .553 & .337 & .457 & .403 & .513 & .242 & .386 & .534 & .551 & .436 & .466 \\
African American Vernacular & .348 & .552 & .325 & .441 & .376 & .441 & .269 & .362 & .539 & .560 & .438 & .491 \\
Southeast England Dialect & .348 & .551 & .328 & .445 & .433 & .455 & .245 & .372 & .548 & .580 & .436 & .477 \\
Rural African American Vernacular & .344 & .550 & .343 & .463 & .349 & .463 & .260 & .332 & .510 & .549 & .436 & .483 \\
Maltese Dialect & .342 & .549 & .343 & .463 & .348 & .492 & .242 & .352 & .525 & .548 & .446 & .480 \\
Irish Dialect & .337 & .547 & .335 & .454 & .368 & .502 & .222 & .368 & .542 & .529 & .403 & .483 \\
Jamaican Dialect & .332 & .545 & .332 & .450 & .370 & .469 & .268 & .360 & .547 & .547 & .429 & .468 \\
Aboriginal Dialect & .329 & .543 & .338 & .458 & .325 & .448 & .231 & .357 & .529 & .517 & .437 & .472 \\
North England Dialect & .328 & .541 & .325 & .442 & .379 & .467 & .234 & .369 & .550 & .565 & .427 & .454 \\
St Helena Dialect & .322 & .539 & .349 & .472 & .382 & .506 & .249 & .360 & .536 & .539 & .426 & .472 \\
Welsh Dialect & .312 & .537 & .331 & .449 & .356 & .485 & .237 & .393 & .532 & .556 & .422 & .492 \\
Southwest England Dialect & .301 & .535 & .323 & .436 & .336 & .446 & .289 & .370 & .512 & .541 & .422 & .477 \\
\bottomrule
\end{tabular}
\end{adjustbox}
\end{table*}

\begin{table*}[t]
    \centering
    \small 
    \caption{MultiAgent Framework Results for GPT-4 on PrivacyQA and PolicyQA}\label{tab:t2}
    \renewcommand{\arraystretch}{1.2} 
    \setlength{\tabcolsep}{6pt} 
    \begin{adjustbox}{max width=\textwidth} 
    \begin{tabular}{lcc|cc|cc|cc}
        \toprule
        \textbf{Dialect} & \multicolumn{2}{c|}{\textbf{PrivacyQA Zero-shot}} & \multicolumn{2}{c|}{\textbf{PrivacyQA Few-shot}} & \multicolumn{2}{c|}{\textbf{PolicyQA Zero-shot}} & \multicolumn{2}{c}{\textbf{PolicyQA Few-shot}} \\
        & Initial & Final & Initial & Final & Initial & Final & Initial & Final \\
        \midrule
        Standard American Dialect & .532 & .610 & .608 & .611 & .444 & .464 & .481 & .484 \\
        Tanzanian Dialect & .531 & .586 & .580 & .588 & .437 & .457 & .478 & .481 \\
        Manx Dialect & .531 & .581 & .572 & .600 & .442 & .460 & .478 & .481 \\
        Orkney Shetland Dialect & .527 & .579 & .574 & .602 & .442 & .457 & .474 & .478 \\
        New Zealand Dialect & .527 & .576 & .576 & .598 & .440 & .461 & .474 & .478 \\
        Nigerian Dialect & .532 & .588 & .573 & .600 & .441 & .462 & .474 & .478 \\
        East Anglican Dialect & .528 & .587 & .578 & .604 & .427 & .455 & .474 & .478 \\
        African American Vernacular & .529 & .570 & .577 & .598 & .424 & .460 & .473 & .477 \\
        Early African American Vernacular & .527 & .583 & .577 & .587 & .433 & .459 & .472 & .476 \\
        Black South African Dialect & .533 & .594 & .577 & .598 & .421 & .451 & .471 & .475 \\
        Jamaican Dialect & .529 & .578 & .577 & .596 & .426 & .451 & .471 & .475 \\
        Newfoundland Dialect & .528 & .581 & .576 & .600 & .436 & .452 & .471 & .475 \\
        Australian Vernacular & .528 & .604 & .579 & .601 & .423 & .454 & .470 & .475 \\
        Irish Dialect & .526 & .575 & .577 & .589 & .433 & .450 & .470 & .474 \\
        Fiji Basilect & .525 & .586 & .576 & .596 & .427 & .451 & .469 & .474 \\
        North England Dialect & .525 & .584 & .579 & .601 & .437 & .450 & .469 & .474 \\
        Scottish Dialect & .529 & .580 & .576 & .602 & .427 & .456 & .469 & .474 \\
        St Helena Dialect & .529 & .597 & .581 & .602 & .425 & .449 & .468 & .473 \\
        Aboriginal Dialect & .528 & .587 & .581 & .602 & .418 & .458 & .468 & .473 \\
        Pakistani Dialect & .529 & .597 & .576 & .597 & .420 & .451 & .468 & .472 \\
        Malaysian Dialect & .529 & .581 & .576 & .598 & .436 & .449 & .468 & .472 \\
        Ghanaian Dialect & .529 & .590 & .576 & .597 & .428 & .454 & .468 & .472 \\
        Southeast England Dialect & .526 & .585 & .577 & .595 & .433 & .451 & .468 & .472 \\
        Bahamian Dialect & .530 & .578 & .576 & .596 & .420 & .450 & .467 & .472 \\
        Colloquial Singapore Dialect & .526 & .573 & .578 & .599 & .421 & .454 & .467 & .472 \\
        Falkland Islands Dialect & .529 & .585 & .578 & .592 & .419 & .454 & .467 & .472 \\
        Southeast American Enclave Dialect & .532 & .588 & .576 & .587 & .435 & .455 & .467 & .471 \\
        Welsh Dialect & .529 & .592 & .577 & .592 & .433 & .447 & .465 & .469 \\
        Australian Dialect & .531 & .582 & .574 & .602 & .435 & .449 & .465 & .469 \\
        White Zimbabwean Dialect & .528 & .590 & .574 & .597 & .425 & .451 & .464 & .469 \\
        Ozark Dialect & .530 & .589 & .578 & .597 & .423 & .451 & .464 & .469 \\
        Channel Islands Dialect & .530 & .584 & .579 & .589 & .425 & .450 & .463 & .468 \\
        Chicano Dialect & .530 & .604 & .582 & .611 & .419 & .445 & .463 & .468 \\
        Cape Flats Dialect & .528 & .581 & .577 & .590 & .421 & .447 & .463 & .468 \\
        Colloquial American Dialect & .528 & .577 & .578 & .600 & .421 & .447 & .463 & .468 \\
        Kenyan Dialect & .525 & .593 & .582 & .592 & .415 & .449 & .462 & .467 \\
        White South African Dialect & .529 & .588 & .577 & .604 & .430 & .444 & .462 & .467 \\
        Ugandan Dialect & .532 & .601 & .580 & .590 & .421 & .444 & .462 & .467 \\
        Southwest England Dialect & .527 & .576 & .581 & .594 & .415 & .445 & .462 & .467 \\
        Appalachian Dialect & .527 & .589 & .575 & .595 & .416 & .449 & .461 & .466 \\
        Tristan Dialect & .526 & .584 & .575 & .592 & .429 & .443 & .460 & .465 \\
        Indian Dialect & .531 & .585 & .577 & .600 & .414 & .443 & .459 & .465 \\
        Cameroon Dialect & .527 & .590 & .580 & .585 & .420 & .440 & .458 & .463 \\
        Hong Kong Dialect & .528 & .594 & .577 & .601 & .410 & .439 & .458 & .463 \\
        Indian South African Dialect & .527 & .590 & .577 & .596 & .415 & .444 & .457 & .463 \\
        Rural African American Vernacular & .527 & .588 & .573 & .595 & .424 & .444 & .454 & .460 \\
        Maltese Dialect & .529 & .592 & .576 & .597 & .408 & .441 & .454 & .460 \\
        \bottomrule
    \end{tabular}
    \end{adjustbox}
\end{table*}

\begin{table*}[t]
    \centering
    \small
    \caption{MultiAgent Framework Results for Llama 3.1 on PrivacyQA and PolicyQA}\label{tab:t3}
    \renewcommand{\arraystretch}{1.2} 
    \setlength{\tabcolsep}{6pt} 
    \begin{adjustbox}{max width=\textwidth} 
    \begin{tabular}{lcc|cc|cc|cc}
        \toprule
        \textbf{Dialect} & \multicolumn{2}{c|}{\textbf{PrivacyQA Zero-shot}} & \multicolumn{2}{c|}{\textbf{PrivacyQA Few-shot}} & \multicolumn{2}{c|}{\textbf{PolicyQA Zero-shot}} & \multicolumn{2}{c}{\textbf{PolicyQA Few-shot}} \\
        & Initial & Final & Initial & Final & Initial & Final & Initial & Final \\
        \midrule
        Standard American Dialect & .514 & .549 & .424 & .555 & .310 & .381 & .379 & .400 \\
        St Helena Dialect & .493 & .514 & .488 & .536 & .241 & .368 & .335 & .392 \\
        Kenyan Dialect & .506 & .559 & .502 & .543 & .264 & .355 & .352 & .361 \\
        Scottish Dialect & .508 & .535 & .510 & .545 & .260 & .360 & .350 & .385 \\
        Ozark Dialect & .498 & .519 & .505 & .552 & .268 & .382 & .357 & .372 \\
        New Zealand Dialect & .493 & .512 & .480 & .503 & .228 & .352 & .317 & .384 \\
        Ugandan Dialect & .502 & .533 & .507 & .549 & .242 & .351 & .332 & .374 \\
        Early African American Vernacular & .505 & .540 & .510 & .523 & .257 & .374 & .344 & .387 \\
        Indian South African Dialect & .495 & .546 & .501 & .519 & .231 & .372 & .326 & .374 \\
        Falkland Islands Dialect & .495 & .511 & .496 & .524 & .246 & .374 & .342 & .387 \\
        Colloquial Singapore Dialect & .514 & .527 & .501 & .513 & .251 & .366 & .344 & .377 \\
        Welsh Dialect & .497 & .523 & .491 & .522 & .290 & .372 & .371 & .394 \\
        Indian Dialect & .496 & .536 & .492 & .509 & .210 & .377 & .310 & .397 \\
        Malaysian Dialect & .506 & .529 & .497 & .532 & .244 & .386 & .345 & .364 \\
        Irish Dialect & .497 & .521 & .494 & .507 & .248 & .376 & .346 & .377 \\
        White Zimbabwean Dialect & .513 & .537 & .501 & .536 & .237 & .358 & .328 & .390 \\
        African American Vernacular & .488 & .527 & .502 & .525 & .242 & .374 & .338 & .385 \\
        Tristan Dialect & .510 & .521 & .511 & .534 & .208 & .369 & .314 & .382 \\
        Jamaican Dialect & .492 & .520 & .476 & .523 & .240 & .368 & .324 & .391 \\
        Newfoundland Dialect & .512 & .545 & .521 & .539 & .260 & .355 & .352 & .389 \\
        White South African Dialect & .532 & .539 & .518 & .522 & .285 & .380 & .358 & .379 \\
        Appalachian Dialect & .501 & .532 & .497 & .529 & .246 & .360 & .330 & .386 \\
        Ghanaian Dialect & .517 & .549 & .512 & .542 & .239 & .364 & .319 & .381 \\
        Australian Vernacular & .501 & .528 & .497 & .524 & .289 & .372 & .366 & .388 \\
        Channel Islands Dialect & .529 & .550 & .527 & .548 & .263 & .369 & .342 & .369 \\
        Hong Kong Dialect & .507 & .525 & .485 & .520 & .222 & .364 & .311 & .396 \\
        Black South African Dialect & .507 & .530 & .483 & .516 & .245 & .374 & .345 & .366 \\
        Maltese Dialect & .534 & .564 & .499 & .525 & .231 & .381 & .329 & .374 \\
        Rural African American Vernacular & .489 & .523 & .496 & .538 & .207 & .374 & .309 & .380 \\
        Southeast England Dialect & .530 & .548 & .514 & .536 & .257 & .370 & .341 & .374 \\
        Pakistani Dialect & .522 & .560 & .514 & .536 & .240 & .351 & .334 & .387 \\
        Fiji Acrolect & .502 & .530 & .494 & .534 & .270 & .373 & .348 & .372 \\
        Southeast American Enclave Dialect & .497 & .520 & .500 & .527 & .250 & .357 & .337 & .377 \\
        East Anglican Dialect & .487 & .502 & .480 & .510 & .260 & .356 & .340 & .388 \\
        Orkney Shetland Dialect & .513 & .540 & .515 & .520 & .265 & .351 & .350 & .370 \\
        Bahamian Dialect & .503 & .521 & .488 & .510 & .234 & .368 & .329 & .368 \\
        Manx Dialect & .486 & .506 & .489 & .532 & .287 & .383 & .355 & .377 \\
        Cameroon Dialect & .521 & .545 & .481 & .511 & .228 & .374 & .324 & .388 \\
        North England Dialect & .518 & .539 & .495 & .525 & .249 & .369 & .337 & .377 \\
        Colloquial American Dialect & .496 & .520 & .506 & .530 & .241 & .384 & .329 & .394 \\
        Australian Dialect & .506 & .537 & .488 & .519 & .237 & .359 & .323 & .389 \\
        Fiji Basilect & .491 & .536 & .468 & .505 & .269 & .381 & .344 & .374 \\
        Nigerian Dialect & .498 & .547 & .495 & .524 & .262 & .377 & .345 & .368 \\
        Philippine Dialect & .505 & .537 & .498 & .515 & .247 & .361 & .341 & .387 \\
        Sri Lankan Dialect & .530 & .556 & .512 & .544 & .256 & .373 & .348 & .373 \\
        Liberian Settler Dialect & .507 & .531 & .492 & .509 & .264 & .381 & .356 & .381 \\
        Tanzanian Dialect & .517 & .542 & .498 & .533 & .276 & .377 & .346 & .371 \\
        Cape Flats Dialect & .511 & .521 & .514 & .544 & .268 & .378 & .358 & .374 \\
        \bottomrule
    \end{tabular}
    \end{adjustbox}
\end{table*}

\begin{table*}[t]
    \centering
    \small
    \caption{MultiAgent Framework Results for DeepSeek-R1 on PrivacyQA and PolicyQA} \label{tab:t4}
    \renewcommand{\arraystretch}{1.2} 
    \setlength{\tabcolsep}{6pt} 
    \begin{adjustbox}{max width=\textwidth} 
    \begin{tabular}{lcc|cc|cc|cc}
        \toprule
        \textbf{Dialect} & \multicolumn{2}{c|}{\textbf{PrivacyQA Zero-shot}} & \multicolumn{2}{c|}{\textbf{PrivacyQA Few-shot}} & \multicolumn{2}{c|}{\textbf{PolicyQA Zero-shot}} & \multicolumn{2}{c}{\textbf{PolicyQA Few-shot}} \\
        & Initial & Final & Initial & Final & Initial & Final & Initial & Final \\
        \midrule
        Kenyan Dialect & .517 & .569 & .446 & .585 & .446 & .488 & .428 & .498 \\
        St Helena Dialect & .543 & .587 & .456 & .569 & .416 & .468 & .460 & .491 \\
        Scottish Dialect & .529 & .580 & .440 & .535 & .419 & .481 & .464 & .485 \\
        Ozark Dialect & .515 & .575 & .478 & .581 & .404 & .470 & .421 & .498 \\
        New Zealand Dialect & .530 & .583 & .439 & .569 & .406 & .465 & .422 & .494 \\
        Ugandan Dialect & .535 & .578 & .437 & .563 & .401 & .475 & .430 & .477 \\
        Early African American Vernacular & .526 & .584 & .481 & .580 & .401 & .475 & .460 & .491 \\
        Indian South African Dialect & .523 & .578 & .436 & .573 & .411 & .488 & .451 & .473 \\
        Falkland Islands Dialect & .532 & .569 & .440 & .535 & .437 & .480 & .443 & .483 \\
        Colloquial Singapore Dialect & .498 & .570 & .436 & .572 & .417 & .488 & .433 & .495 \\
        Indian Dialect & .532 & .583 & .460 & .584 & .416 & .459 & .455 & .479 \\
        Malaysian Dialect & .501 & .569 & .439 & .552 & .434 & .488 & .461 & .506 \\
        Irish Dialect & .504 & .560 & .445 & .578 & .431 & .468 & .449 & .500 \\
        African American Vernacular & .512 & .551 & .462 & .578 & .436 & .485 & .464 & .490 \\
        Jamaican Dialect & .517 & .583 & .447 & .585 & .437 & .474 & .455 & .494 \\
        Standard American Dialect & .501 & .562 & .460 & .533 & .422 & .451 & .456 & .474 \\
        Newfoundland Dialect & .531 & .575 & .448 & .557 & .437 & .468 & .433 & .475 \\
        Appalachian Dialect & .519 & .560 & .470 & .567 & .414 & .453 & .451 & .488 \\
        Ghanaian Dialect & .514 & .561 & .468 & .564 & .448 & .482 & .424 & .486 \\
        Australian Vernacular & .550 & .602 & .434 & .561 & .434 & .467 & .413 & .481 \\
        Channel Islands Dialect & .507 & .574 & .466 & .554 & .429 & .458 & .428 & .475 \\
        Hong Kong Dialect & .507 & .579 & .448 & .557 & .434 & .485 & .419 & .502 \\
        Black South African Dialect & .515 & .571 & .445 & .590 & .440 & .474 & .420 & .471 \\
        Maltese Dialect & .512 & .576 & .451 & .565 & .440 & .469 & .436 & .504 \\
        Rural African American Vernacular & .534 & .579 & .476 & .606 & .420 & .480 & .467 & .476 \\
        Pakistani Dialect & .507 & .568 & .452 & .579 & .411 & .469 & .422 & .493 \\
        Fiji Acrolect & .551 & .579 & .486 & .573 & .403 & .472 & .451 & .485 \\
        Southeast American Enclave Dialect & .510 & .582 & .463 & .593 & .424 & .452 & .422 & .505 \\
        East Anglican Dialect & .523 & .593 & .459 & .569 & .444 & .467 & .452 & .498 \\
        Orkney Shetland Dialect & .511 & .563 & .456 & .572 & .447 & .456 & .417 & .487 \\
        Bahamian Dialect & .517 & .574 & .449 & .572 & .410 & .470 & .425 & .506 \\
        Manx Dialect & .551 & .580 & .450 & .580 & .445 & .482 & .416 & .471 \\
        Cameroon Dialect & .517 & .575 & .470 & .573 & .432 & .472 & .462 & .484 \\
        North England Dialect & .522 & .577 & .464 & .587 & .448 & .471 & .430 & .504 \\
        Colloquial American Dialect & .530 & .586 & .465 & .577 & .444 & .487 & .454 & .494 \\
        Australian Dialect & .525 & .580 & .465 & .577 & .400 & .467 & .445 & .471 \\
        Fiji Basilect & .532 & .571 & .476 & .605 & .449 & .482 & .414 & .501 \\
        Nigerian Dialect & .522 & .571 & .467 & .551 & .411 & .479 & .468 & .484 \\
        Philippine Dialect & .522 & .580 & .464 & .566 & .422 & .472 & .428 & .477 \\
        Sri Lankan Dialect & .537 & .555 & .468 & .563 & .428 & .459 & .469 & .505 \\
        Liberian Settler Dialect & .547 & .583 & .455 & .572 & .433 & .478 & .469 & .502 \\
        Tanzanian Dialect & .534 & .584 & .456 & .574 & .400 & .486 & .413 & .503 \\
        Cape Flats Dialect & .537 & .596 & .443 & .548 & .434 & .451 & .424 & .478 \\
        \bottomrule
    \end{tabular}
    \end{adjustbox}
\end{table*}

\begin{table*}[t]
    \centering
    \small
    \caption{Few-shot MultiAgent Framework Results for GPT-4o-mini on PrivacyQA(No Dialect Info)}\label{tab:nodialectfull}
    \renewcommand{\arraystretch}{1.2} 
    \setlength{\tabcolsep}{6pt}      
    \begin{adjustbox}{max width=\textwidth}
    \begin{tabular}{lcc}
        \toprule
        \textbf{Dialect} & \textbf{Initial F1} & \textbf{Final F1} \\
        \midrule
StHelenaDialect                & 0.529 & 0.555 \\
KenyanDialect                  & 0.533 & 0.600 \\
ScottishDialect                & 0.521 & 0.603 \\
OzarkDialect                   & 0.518 & 0.583 \\
NewZealandDialect              & 0.516 & 0.600 \\
UgandanDialect                 & 0.535 & 0.597 \\
EarlyAfricanAmericanVernacular & 0.532 & 0.558 \\
IndianSouthAfricanDialect      & 0.516 & 0.590 \\
FalklandIslandsDialect         & 0.527 & 0.564 \\
ColloquialSingaporeDialect     & 0.508 & 0.600 \\
WelshDialect                   & 0.524 & 0.610 \\
IndianDialect                  & 0.518 & 0.578 \\
MalaysianDialect               & 0.510 & 0.565 \\
IrishDialect                   & 0.501 & 0.556 \\
WhiteZimbabweanDialect         & 0.527 & 0.574 \\
AfricanAmericanVernacular      & 0.534 & 0.576 \\
TristanDialect                 & 0.534 & 0.553 \\
JamaicanDialect                & 0.513 & 0.600 \\
StandardAmericanDialect        & 0.518 & 0.614 \\
NewfoundlandDialect            & 0.512 & 0.604 \\
WhiteSouthAfricanDialect       & 0.516 & 0.567 \\
AppalachianDialect             & 0.530 & 0.605 \\
GhanaianDialect                & 0.520 & 0.603 \\
AustralianVernacular           & 0.534 & 0.595 \\
ChannelIslandsDialect          & 0.508 & 0.596 \\
HongKongDialect                & 0.522 & 0.605 \\
BlackSouthAfricanDialect       & 0.512 & 0.564 \\
MalteseDialect                 & 0.496 & 0.606 \\
RuralAfricanAmericanVernacular & 0.501 & 0.604 \\
SoutheastEnglandDialect        & 0.518 & 0.565 \\
PakistaniDialect               & 0.523 & 0.599 \\
FijiAcrolect                   & 0.526 & 0.582 \\
SoutheastAmericanEnclaveDialect & 0.539 & 0.612 \\
EastAnglicanDialect            & 0.514 & 0.591 \\
OrkneyShetlandDialect          & 0.521 & 0.622 \\
BahamianDialect                & 0.508 & 0.592 \\
ManxDialect                    & 0.514 & 0.575 \\
CameroonDialect                & 0.526 & 0.566 \\
NorthEnglandDialect            & 0.531 & 0.565 \\
ColloquialAmericanDialect      & 0.513 & 0.573 \\
AustralianDialect              & 0.526 & 0.587 \\
FijiBasilect                   & 0.536 & 0.619 \\
NigerianDialect                & 0.531 & 0.603 \\
PhilippineDialect              & 0.522 & 0.595 \\
SriLankanDialect               & 0.525 & 0.622 \\
LiberianSettlerDialect         & 0.518 & 0.584 \\
TanzanianDialect               & 0.521 & 0.615 \\
CapeFlatsDialect               & 0.530 & 0.594 \\
        \bottomrule
    \end{tabular}
    \end{adjustbox}
\end{table*}

\end{document}